\documentclass[pdflatex,sn-nature]{sn-jnl}

\usepackage{graphicx}%
\usepackage{multirow}%
\usepackage{amsmath,amssymb,amsfonts}%
\usepackage{amsthm}%
\usepackage{mathrsfs}%
\usepackage[title]{appendix}%
\usepackage{xcolor}%
\usepackage{textcomp}%
\usepackage{manyfoot}%
\usepackage{booktabs}%
\usepackage{algorithm}%
\usepackage{algorithmicx}%
\usepackage{algpseudocode}%
\usepackage{listings}%
\usepackage{tcolorbox}
\tcbuselibrary{breakable, skins}
\usepackage{setspace}
\usepackage{enumitem}   
\usepackage{natbib}

\theoremstyle{thmstyleone}%
\theoremstyle{thmstyletwo}%

\theoremstyle{thmstylethree}%

\tcbuselibrary{skins, breakable}

\newtcolorbox{llmprompt}[1][]{%
    colback=gray!10,      
    colframe=gray!40,     
    coltitle=black,       
    fonttitle=\bfseries,  
    title=System Prompt,  
    breakable,            
    enhanced,
    #1
}

\colorlet{punct}{red!60!black}
\definecolor{background}{HTML}{EEEEEE}
\definecolor{delim}{RGB}{20,105,176}
\colorlet{numb}{magenta!60!black}

\usepackage{listings}
\usepackage{xcolor}
\usepackage{ragged2e} 

\colorlet{punct}{red!60!black}
\definecolor{background}{HTML}{EEEEEE}
\definecolor{delim}{RGB}{20,105,176}
\colorlet{numb}{magenta!60!black}
\lstdefinelanguage{json}{
    basicstyle=\scriptsize\ttfamily,
    numbers=left,
    numberstyle=\scriptsize,
    stepnumber=1,
    numbersep=8pt,
    showstringspaces=false,
    breaklines=true,
    frame=lines,
    backgroundcolor=\color{background},
    literate=
     *{0}{{{\color{numb}0}}}{1}
      {1}{{{\color{numb}1}}}{1}
      {2}{{{\color{numb}2}}}{1}
      {3}{{{\color{numb}3}}}{1}
      {4}{{{\color{numb}4}}}{1}
      {5}{{{\color{numb}5}}}{1}
      {6}{{{\color{numb}6}}}{1}
      {7}{{{\color{numb}7}}}{1}
      {8}{{{\color{numb}8}}}{1}
      {9}{{{\color{numb}9}}}{1}
      {:}{{{\color{punct}{:}}}}{1}
      {,}{{{\color{punct}{,}}}}{1}
      {\{}{{{\color{delim}{\{}}}}{1}
      {\}}{{{\color{delim}{\}}}}}{1}
      {[}{{{\color{delim}{[}}}}{1}
      {]}{{{\color{delim}{]}}}}{1},
}
\usepackage{amsmath,amssymb}

\definecolor{promptbg}{RGB}{248, 249, 250}
\definecolor{promptframe}{RGB}{108, 117, 125}
\definecolor{systembg}{RGB}{232, 240, 254}
\definecolor{systemframe}{RGB}{66, 133, 244}
\definecolor{speakerbg}{RGB}{232, 248, 240}
\definecolor{speakerframe}{RGB}{52, 168, 83}
\definecolor{hearerbg}{RGB}{254, 243, 232}
\definecolor{hearerframe}{RGB}{251, 133, 0}
\definecolor{feedbackbg}{RGB}{253, 232, 232}
\definecolor{feedbackframe}{RGB}{234, 67, 53}
\definecolor{codefont}{RGB}{60, 60, 60}
 
\tcbset{
    promptbox/.style={
        breakable,
        enhanced,
        colback=#1bg,
        colframe=#1frame,
        fonttitle=\sffamily\bfseries\small,
        coltitle=white,
        attach boxed title to top left={yshift=-2mm, xshift=4mm},
        boxed title style={colback=#1frame, colframe=#1frame,
                           rounded corners, size=small},
        arc=4pt,
        boxrule=0.8pt,
        left=6pt, right=6pt, top=8pt, bottom=6pt,
        before upper={\setstretch{1.1}\ttfamily\footnotesize\color{codefont}},
    }
}
 
\begin{document}

\title[Article Title]{Lexical discovery in unknown environments orchestrated by Large Language Models}


\author*[1]{\fnm{Rafael} \sur{Sendra-Arranz}}\email{r.sendra@csic.es}
\author[1]{\fnm{Iñaki} \sur{Dellibarda Varela}}
\author[1]{\fnm{Eduardo} \sur{Rocon}}
\author[2]{\fnm{Álvaro} \sur{Gutiérrez}}
\author[1]{\fnm{Manuel} \sur{Cebrian}}
\equalcont{Deceased 31 March 2026.}
\affil[1]{\orgdiv{Center for Automation and Robotics}, \orgname{Spanish National Research Council (CSIC-UPM)}, \orgaddress{\city{Madrid}, \country{Spain}}}
\affil[2]{\orgdiv{E.T.S. Ingenieros de Telecomunicación}, \orgname{Universidad Politécnica de Madrid}, \orgaddress{\city{Madrid}, \postcode{28040}, \country{Spain}}}


\abstract{Populations of autonomous agents deployed in unknown 
environments (e.g.\ planetary or deep-sea exploration) must 
develop shared vocabularies to refer to entities that have no 
name in any human language. We propose the Neuro-Symbolic Lexical 
Discovery (NSLD) framework, in which a population of LLM-based 
agents plays a referential game over out-of-distribution visual 
referents, autonomously self-organising a shared alien lexicon. 
Each agent combines a frozen CLIP vision encoder with a private 
FAISS vector index and a text-only LLM. Crucially, discovered 
alien words are anchored to natural language via semantic 
proximity in the embedding space, enlarging the human 
vocabulary with new perceptually grounded words. Consensus is 
reached in simulations with populations of up to twenty agents 
and ten visual referents. Convergence dynamics are characterised 
through three analytical models achieving $R^2 > 0.95$, 
representing a first step towards pre-deployment planning in 
autonomous exploration missions.}

\keywords{symbol grounding, emergent communication, large language 
models, vision-language models, referential games, cultural 
transmission, lexical convergence, multi-agent systems}



\maketitle

\section{Introduction}\label{sec:intro}

The rapid ascent of Large Language Models (LLMs) has disrupted the 
landscape of Artificial Intelligence (AI), transitioning from specialized natural 
language processing techniques to general-purpose technologies with implications in fields such as 
chemistry~\cite{ramos25,bran24,zheng25}, medicine~\cite{thirunavukarasu23,liu25,shool25}, 
robotics~\cite{monWilliams25,ahn22,huang22}, search and rescue missions~\cite{jarabo25}, and 
education~\cite{chu25,zhang-etal25}. 
\begin{figure}[t!]
    \centering
    \includegraphics[width=1.0\linewidth]{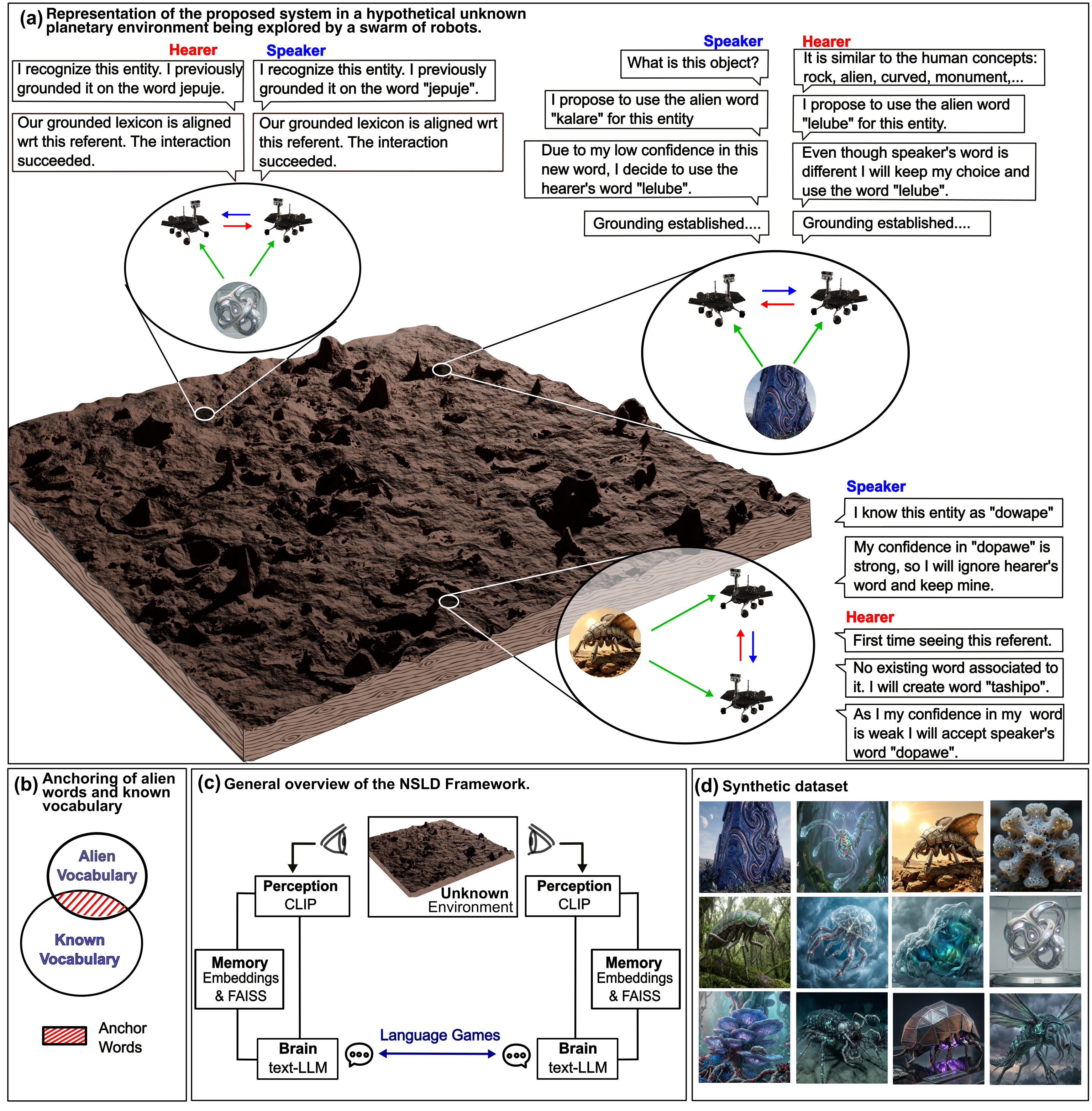}
\caption{\textbf{Overview of the Neuro-Symbolic Lexical Discovery (NSLD) framework.}
(\textbf{a}) Conceptual illustration of the NSLD framework in a hypothetical planetary 
exploration scenario. A swarm of autonomous robots must collectively develop a shared 
grounded vocabulary to refer to unnameable visual entities. Three representative 
interaction outcomes are shown: a successful round with prior grounding (top left); a 
failed round resolved by confidence-driven lexical revision (top right); and a 
first-encounter round in which the hearer defers to the speaker's more confident choice 
(bottom right). The planetary environment and robot swarm are depicted for illustrative 
purposes only and are not simulated in this work. Robot 3D models being depicted in the figure 
are reproduced from the NASA 3D Resources repository~\cite{nasa3d}. 
(\textbf{b}) Alien words are anchored to natural language via semantic proximity in the 
shared embedding space.
(\textbf{c}) General architecture of the NSLD framework, in which pairs of agents 
interact via the Grounding Referential Game.
(\textbf{d}) Synthetic dataset of OOD visual referents used in all experiments, 
generated using \textit{Nano Banana}. Although AI-generated, these images constitute 
the actual experimental dataset.}
\label{fig:intro}
\end{figure}

LLM agents are increasingly deployed in multi-agent systems where 
coordination, communication, and collective planning are of essence~\cite{park23,guo24}. 
Yet a fundamental bottleneck arises when agents operate in never-explored environments and encounter entities with no name 
in any human language. Consider a swarm of autonomous robots surveying the subsurface ocean of Europa (a Jupiter's icy moon): 
upon encountering an intricate mineral formation unlike anything reported on Earth, the robots must collectively develop a 
shared grounded vocabulary to refer to it, and associate those words with natural language descriptors to report 
their findings to human operators.

Recent work has demonstrated that LLM-controlled robots can complete 
complex long-horizon tasks in unpredictable and diverse physical environments~\cite{monWilliams25,ahn22,huang22}. 
However, these frameworks assume that all encountered entities are nameable in natural language, which is 
a condition that breaks down in genuinely unknown environments without prior linguistic representations.
Even though LLMs excel at communicating through natural language, 
they lack mechanisms to ground novel symbols on observable referents that are    
unnameable in natural language, nor is there any procedure to associate newly acquired 
\emph{alien}  (i.e., absent from any natural language) 
symbols with known human words. The challenge of developing grounded shared vocabularies between 
humans and machines has been identified as 
a largely open problem~\cite{kouwenhoven2022}, with existing 
approaches either producing semantically opaque tokens~\cite{FlintAshery25} 
or requiring direct human participation in the vocabulary construction process~\cite{steels1999}. Far from being a philosophical claim, we highlight this as a 
concrete engineering bottleneck that must be carefully addressed for the effective deployment 
of LLMs in robotics. 

The Symbol Grounding Problem (SGP)~\cite{harnad90} provides the 
theoretical frame for this bottleneck. It states that symbols used by agents cannot be abstract but must be grounded on referents in the real-world. 
Many authors agree that the symbols manipulated by LLMs lack such grounding, primarily because LLMs 
acquire their world models indirectly through language corpora, without any sensorimotor 
interaction with the world~\cite{harnad25,bender20}. 
The \emph{Collective World Model hypothesis}~\cite{taniguchi24a,taniguchi26} argues that LLMs 
inherit indirect grounding through natural language 
corpora encoding the aggregated sensorimotor experience of human speakers. 
Empirical evidence confirms this deficit: visual training 
partially closes the sensorimotor gap but does not resolve it~\cite{xu25}.

Rather than arguing against this claim and investigating whether LLMs already possess human-like 
grounding on known natural objects~\cite{du25}, we ask how a 
population of LLMs can acquire visual grounding on alien
referents that are entirely absent from natural language. We accomplish this study by coupling 
perceptual experience to persistent symbolic memory and collective negotiation. 

To this end, we propose the Neuro-Symbolic Lexical Discovery (NSLD) 
framework, in which a population of LLM-based agents, deployed, for instance, in robot swarms exploring a hypothetical unknown planet 
(Fig.~\ref{fig:intro}a), play a referential game over a set of 
out-of-distribution (OOD) visual referents. Each agent combines frozen CLIP encoders~\cite{radford2021clip} (vision module), a 
private FAISS vector index~\cite{johnson2019faiss} (grounded memory), and text-only LLMs (reasoning and 
lexical decision-making), as illustrated in Fig.~\ref{fig:intro}c. Rounds of the 
grounding referential game are played between pairs of agents, a \emph{speaker} and a \emph{hearer}, 
that independently observe the same referent and must converge on a shared alien word to name it 
(Fig.~\ref{fig:intro}a). Grounding is encoded persistently in an embedding space rather 
than in conversational context, enabling stable symbol--referent associations to emerge 
and diffuse across the population through cultural transmission. Crucially, discovered 
alien words are not isolated symbols: agents anchor them to the natural language 
vocabulary via semantic proximity in the shared embedding space 
(Fig.~\ref{fig:intro}b), effectively enlarging human vocabulary with new 
perceptually grounded terms. The synthetic dataset of OOD visual referents used in 
all experiments is shown in Fig.~\ref{fig:intro}d. 

We demonstrate empirically that populations 
of agents reliably achieve convergence to grounding consensus 
on OOD visual referents across all tested configurations (spanning populations up to twenty agents and up to ten visual referents). 
Not only do agents reach a shared alien vocabulary but they also 
associate it to known words in natural language, creating 
semantically grounded alien-to-English links that support human--machine interaction. 
Moreover, we characterize the dynamics of lexical convergence, proposing three analytical models: (i) a \emph{direct model} that predicts the degree of 
grounding consensus at any round in the referential game, (ii) the \emph{inverse model} that estimates the number of rounds required to reach a certain degree of consensus, 
and (iii) the \emph{vocabulary model} that estimates the evolution of the global vocabulary size. All three models achieve $R^2 > 0.95$, providing accurate 
predictions of lexical convergence dynamics and enabling deployment planning prior to real-world exploration missions.

\subsection*{Related Work}
The NSLD framework is in the intersection of three axes: classical emergent communication and naming games, deep 
learning-based emergent communication, and multi-agent LLM coordination. 

Classical naming games~\cite{steels95,baronchelli06} 
demonstrate that populations of simple agents can self-organize shared vocabularies through pairwise interactions, 
proving consensus convergence via \emph{winner-take-all} dynamics~\cite{baronchelli06}. 
However, these agents acquire semantically isolated labels: words are treated as abstract tokens, that remain ungrounded to physical referents and lack attachment to sensorimotor experience. 
Multiple authors have implemented 
cultural evolution of language in embodied robots~\cite{steels98,steels99,baronchelli06,depaula15,steels12,steels15}. 
One of the pioneering studies in this field is the \emph{talking heads} experiment~\cite{steels1999,steels15}: agents played 
language games whilst embodied in two pan-tilted cameras observing multiple geometric shapes of varying sizes and colours. 
Although agents could not interact with the environment, they were embodied and situated, grounding their vocabulary in direct 
perceptual experience.



In the LLM era, Ashery et al.~\cite{FlintAshery25} demonstrated that populations of LLM agents can autonomously converge to a shared set of words through local pairwise interactions. 
They showed that several collective biases emerge even when individual agents are unbiased. 
However, their framework operates on abstract symbols with no perceptual grounding, perception, and no connection to natural language. 

In~\cite{kouwenhoven25}, the authors explored the emergence of structure in an arbitrary vocabulary that was initially unstructured. 
Two LLMs played traditional referential games with the goal of naming objects described by properties such as color, shape, and amount. 
However, agents lacked perceptual grounding, as referents were described by natural language text properties rather than visual perception, and were drawn from 
a known structured meaning space rather than OOD visual referents.

The \emph{Collective Predictive Coding} (CPC) theoretical framework proposed by Taniguchi and 
colleagues~\cite{taniguchi24a,taniguchi26} is closely related  to our work. CPC interprets symbol emergence in multi-agent systems 
as a decentralised Bayesian inference process in which agents minimise the collective free energy of the system. Agents 
operate in a micro-macro loop: (i) at the micro-level, internal latent representations are learned by each agent 
independently, and (ii) at the macro-level, agents negotiate the emerged symbol system without explicit 
exchange of private latent representations. 
Taniguchi et al. first instantiated this framework using deep generative models~\cite{taniguchi23} and subsequently extended 
it to support the hypothesis that LLMs inherit indirect grounding from natural language corpora~\cite{taniguchi26};
A recent empirical study demonstrates that decentralised multi-agent world models can 
achieve both symbol emergence and coordinated behavior simultaneously~\cite{nomura25}, yet still operating on known 
environments with bridge to natural language. Although CPC shares clear 
similarities with NSLD (e.g. latent representations, referential games, and population dynamics) the NSLD framework addresses a fundamentally non-overlapping problem: rather than 
assuming that LLMs inherit grounding indirectly, we ask whether LLM agents can acquire novel grounding natively and 
associate it with an existing symbol system.


\section{Results}\label{sec::results}
\subsection{Neuro-Symbolic Lexical Discovery Framework}
\label{sec:results:nsld-overview}

The NSLD framework endows LLM-based agents with the ability to perceive, ground, and negotiate novel alien words to refer to OOD 
visual referents, whilst enlarging their natural language vocabulary by establishing semantic associations between 
pre-existing words and discovered alien terms. 
The framework is neuro-symbolic in the sense that it combines (a) 
generative neural networks for vision, reasoning, communication, and decision making, and (b) symbolic associations that create persistent links between 
referents, meanings, and words (the \emph{semiotic triad}~\cite{chandler07}) which are essential for the emergence of stable grounding. 
The NSLD framework consists of a population of agents, each composed by the following three components: 
Each agent combines a frozen CLIP encoder that projects visual input into a shared embedding space (perception), 
a private FAISS vector index (lexical memory), and a text-only LLM (reasoning and lexical decisions).

The NSLD framework operates over a population of agents that not only develop their own alien vocabulary grounded on OOD 
entities, but also engage in collective distributed negotiations to converge on a shared vocabulary. This 
cultural evolution is implemented via a referential game, which we define as the \emph{grounding referential game}. 
It is a turn-based iterative process that seeks to maximise the global 
grounding consensus of the population. Each round is played by two agents, a \emph{speaker} and a \emph{hearer}, both selected randomly from the 
population, and uses the synthetic image dataset shown in Fig.~\ref{fig:intro}d as the OOD visual referents.     
A round is composed of the following phases:

\begin{itemize}
    \item[(i)] \emph{Speaker phase}: the speaker observes a 
    set of images containing OOD entities and selects one. It must then 
    produce an alien word to refer to the selected 
    entity (see Methods), correctly grounded according to its 
    Perception--Memory--Brain pipeline.
    \item[(ii)] \emph{Hearer phase}: the hearer receives the same image 
    selected by the speaker (\emph{joint attention}), and 
    must select an alien word from 
    its own vocabulary to refer to it.
    \item[(iii)] \emph{Feedback phase}: the round outcome (successful if both words match) is shared with both agents, who independently update their lexicons guided by private confidence scores (see Feedback and Confidence Dynamics in Methods).
\end{itemize}

The dataset of visual referents (Fig.~\ref{fig:intro}d) consists of synthetically generated images of alien creatures, rocks and artifacts 
with complex textures and morphologies (see Sec.~\ref{sec:methods:game:dataset}), selected to be OOD with respect to the training corpora of both CLIP and the LLM backend. 
Therefore, it is ensured that grounding cannot be biased by pre-existing linguistic or visual knowledge. 
All simulations use \texttt{gpt-oss:20b} as the LLM backend, a 20-billion parameter open-source language model deployed locally 
(see details in Sec.~\ref{sec:methods:reasoning:llm}).

\subsection{Lexical Emergence in a Five-Agent Population}\label{sec:results:5agents}
\begin{figure*}[t!]
    \centering
    \includegraphics[width=\textwidth]{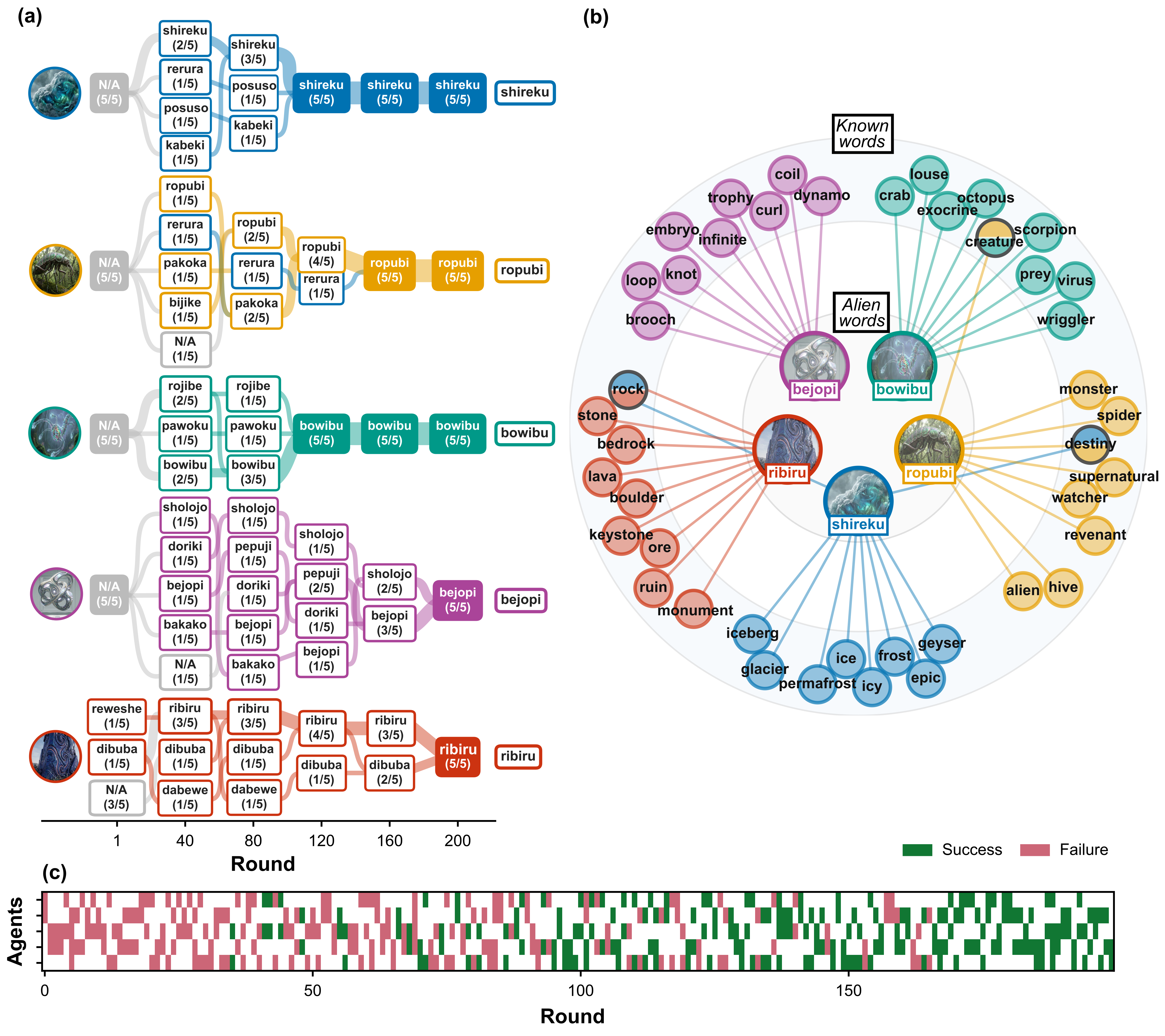}
    \caption{\textbf{Results for a five-agent case study.} 
    (\textbf{a}) Lexicon evolution flow charts tracking the alien words used 
    by the population to refer to each of the five OOD visual referents 
    across a single 200-round trial. Each node displays the word and the 
    number of agents using it at that round. Transient words appear and 
    disappear as the population negotiates, whilst dominant words 
    progressively persist through a winner-take-all dynamics. 
    At the end of the simulation there is a unique shared alien word per referent. 
    (\textbf{b}) Connection between alien words and natural language vocabulary. Each 
    alien word (inner ring) is connected to the OOD visual referent it 
    names (depicted in centre) and to the English descriptors most semantically 
    proximal to it in the shared CLIP embedding space (outer ring). 
    (\textbf{c}) Round-by-round success and failure raster plot for each of 
    the five agents.}
    \label{fig:results-5x5}
\end{figure*}

We present a detailed case study of the lexical emergence and grounding consensus in a population of 
five agents and  five OOD visual entities (referents) over 200 game rounds (Fig.~\ref{fig:results-5x5}).

The lexical dynamics in Fig.~\ref{fig:results-5x5}a reveal that the alien 
vocabulary adopted by the population varies noticeably throughout a grounding 
referential game. Early game dynamics show the creation of many unique 
words,  
with no consensus at all. Subsequently, the number of words in the population 
decreases as rounds of the game increase. Some 
words appear briefly and disappear entirely after some rounds (\emph{transient 
words}), whilst other words persist for longer periods, eventually dominating through a \emph{winner-take-all} dynamic. 
Late game dynamics depict a perfect consensus, concluding with 
a unique common alien word for each entity. 

Fig.~\ref{fig:results-5x5}b demonstrates the integration of the novel alien 
words into the English vocabulary. Not only are new words linked to English 
descriptors as abstract symbols, but they are also grounded to perceptual 
referents. The evolved vocabulary inherits meaningful semantic structure from 
the visual embedding space, as confirmed by the coherent and visually 
interpretable English descriptors associated with the alien words. Furthermore, 
there is minimal overlap in the English 
descriptors of each image, suggesting that grounding is not arbitrary and CLIP 
is able to discriminate among entities. The few cases of descriptor overlap 
(e.g. \texttt{rock} or \texttt{creature}) are consistent with the visual 
similarity of these referents to such concepts.

Fig.~\ref{fig:results-5x5}c depicts the evolution of the game outcome 
(\texttt{SUCCESS} or \texttt{FAILURE}). Early game rounds 
are dominated by failures and  disagreement. In fact, it is not 
until round 35 that agents 3 and 5 record the first successful interaction. 
From this point on, the success rate increases progressively until round 165, 
after which no further failures are recorded.

\subsection{Scaling Laws of Lexical Convergence}\label{sec:results:scaling-laws}
\begin{figure}[t!]
    \centering
    \hspace*{-2cm}
    \includegraphics[width=160mm]{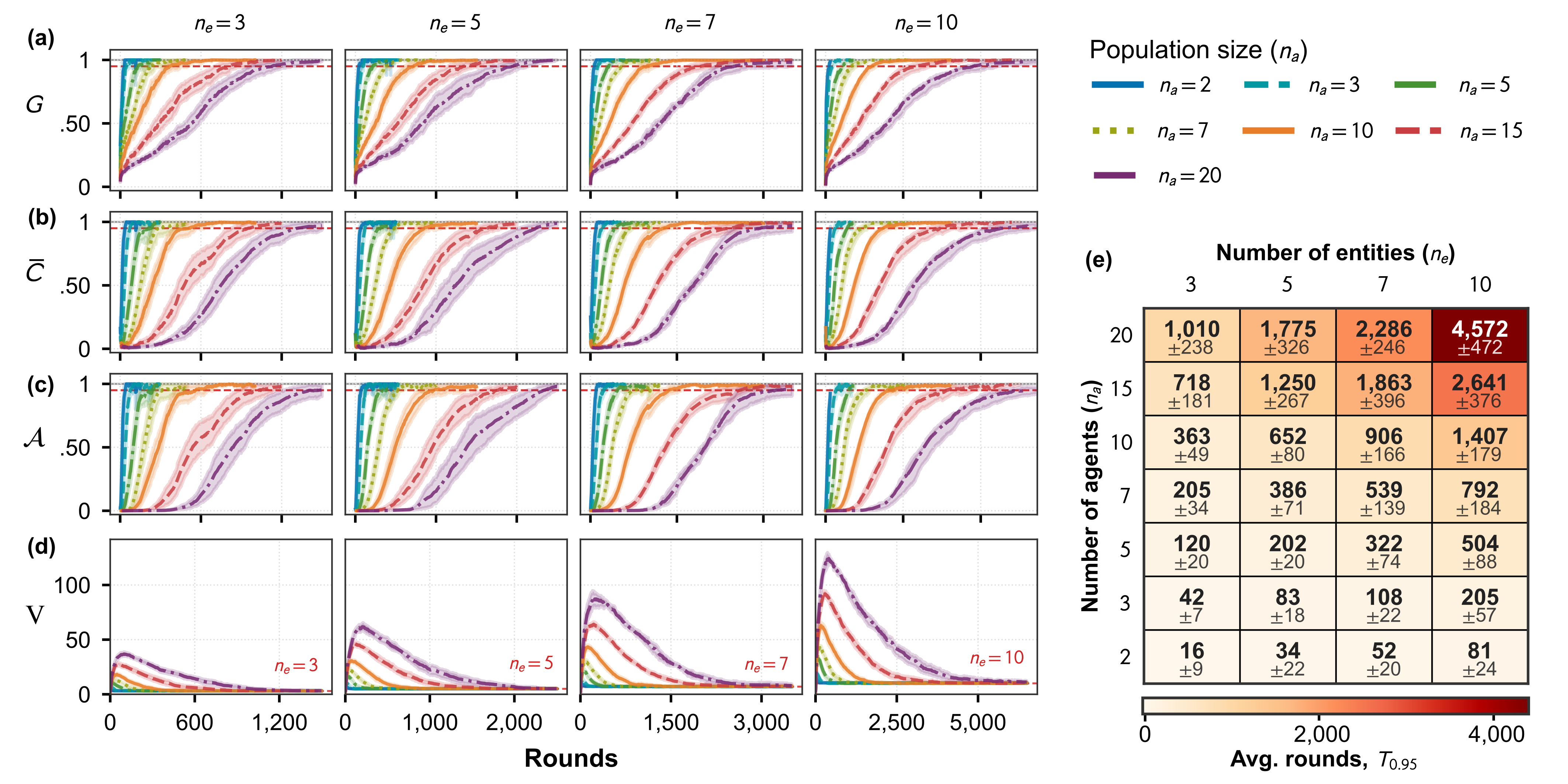}
    \caption{\textbf{Scaling laws of lexical convergence.} (a) Evolution of 
    grounding consensus $G$. (b) Mean population confidence $\bar{C}$, (c) 
    Agreement value $\mathcal{A}$, and (d) global vocabulary size $V$
    across game rounds for all combinations of $n_a \in \{2, 3, 5, 7, 10, 
    15, 20\}$ and $n_e \in \{3, 5, 7, 10\}$. Solid lines show the 
    trial-averaged metric; shaded regions indicate 95\% confidence intervals 
    over 20 independent trials per condition. The horizontal red dashed  lines indicate 
    a 95\% consensus ($G=0.95$), confidence ($\overline{C}=0.95$), and agreement ($\mathcal{A}=0.95$), respectively.  
    (e) Average round $T_{0.95}$ at which populations reach $G(T_{0.95})\approx0.95$ for each combination of $n_a$ and 
    $n_e$, estimated over 20 independent trials per scenario. Values denote 
    mean $\pm$ s.d.}
    \label{fig:joint-plot}
\end{figure}

Fig.~\ref{fig:joint-plot}a-d aggregates the evolution of the proposed metrics (see Methods) throughout 
rounds for different combinations of the population size $n_a$ and the number 
of entities $n_e$. Each subplot displays the average value of the given metric 
and its 95\% Confidence Interval (CI), considering samples of 20 trials per case. 

Fig.~\ref{fig:joint-plot}a reveals that  consensus $G\geq 0.95$ is reached in all the ($n_a$, $n_e$) configurations. It also 
shows that larger values of $n_a$ systematically delay convergence for any 
of the $n_e$ values under consideration. Furthermore, increasing the number 
of entities ($n_e$) also raises convergence time, as can be observed in 
the stretching of the $x$-axis scales as $n_e$ grows. The grounding consensus 
trajectories can be analytically modeled using a logistic model (see Sec.~\ref{sec:results:models}).

The mean confidence $\overline{C}$ and agreement value $\mathcal{A}$ (Figs.~\ref{fig:joint-plot}b--c) both display 
delayed sigmoidal trajectories relative to $G$, with $\mathcal{A}$ additionally exhibiting a zero plateau during 
the early game whose duration scales with $n_a$ and $n_e$.

Moreover, Fig.~\ref{fig:joint-plot}d illustrates the evolution 
of the total number of unique words in the population ($V$). All 
simulations present a rapid explosion of new words in agent vocabularies, 
initially diverging  due to insufficient negotiation. 
The vocabulary size eventually collapses, producing an 
overshoot of $V(T)$ followed by a slower pruning of lexicons 
driven by population dynamics.  Both the overshoot peak value 
and timing are strongly dependent on $n_a$ and $n_e$. In the final rounds, 
the global vocabulary converges to the optimal value of $V=n_e$ in most simulations. 
Even though the steady-state behaviour is not visible in Fig.~\ref{fig:joint-plot}d due to the $x-$axis scale, 
it is detailed in Supplementary Fig.~\ref{fig:supp:zoom}.

Finally, Fig.~\ref{fig:joint-plot}e provides a quantitative summary of the 
results, collecting the 
average round $T_{0.95}$ at which populations reach $G(T_{0.95}) = 0.95$. 
As suggested by Fig.~\ref{fig:joint-plot}a, convergence time grows 
monotonically as both $n_a$ and $n_e$ increase, suggesting a joint multiplicative influence 
on $T_{0.95}$ (see Sec.~\ref{sec:results:models}). 
The fastest convergence is obtained on average in the experiment with $n_a=2$ 
and $n_e=3$, converging in just $16\pm 9$ rounds. In contrast, the slowest 
configuration (twenty agents and ten entities) requires $T_{0.95} = 4572 
\pm 472$ rounds on average, roughly $285\times$ slower. Furthermore, the 
variance of $T_{0.95}$ also scales with both $n_a$ and $n_e$, indicating that 
larger populations with more visual referents yield less predictable convergence times.

\subsection{Mathematical Characterization of Lexical Convergence}\label{sec:results:models}
We present a mathematical model that outputs an estimation of the grounding consensus $G$ at some given round $T$ for any  
population of $n_a$ agents and $n_e$ entities. Such model is logistic,  defined in Eq.~\ref{eq:gss-model},
\begin{equation}\label{eq:gss-model}
    G(T, n_a, n_e) = \dfrac{1}{1+\mathrm{exp}(-k\,(T-T_{0.5}))},
\end{equation}
\noindent where $T_{0.5}$ is the estimated round when $G\approx0.5$ and $k$ is a variable that settles the steepness of the logistic 
curve. Considering that both $T_{0.5}$ and $k$  are visibly dependent on $n_a$ and $n_e$ (see Fig.~\ref{fig:joint-plot}a),
we model them as the bivariate power laws in Eq.~\ref{eq:gss-model:params},
\begin{equation}\label{eq:gss-model:params}
    \left. \begin{array}{r}
         k= c_k \cdot n_a^{\alpha_k}\cdot n_e^{\beta_k}\\
         \\
         T_{0.5} = c_m \cdot n_a^{\alpha_m}\cdot n_e^{\beta_m}
    \end{array}\right\}
\end{equation}
The parameters $(c_k, \alpha_k, \beta_k, c_m, \alpha_m, \beta_m )$ of this model were obtained using nonlinear least squares and 
the Trust Region Reflective algorithm (see Methods).
The choice of the logistic model in Eq.~\ref{eq:gss-model} is motivated by the sigmoid shapes shown in
Fig.~\ref{fig:joint-plot}a. The derivation of the direct model and its power law parameters are detailed in 
Sec.~\ref{sec:methods:models}.

From the $G$ model in Eq.~\ref{eq:gss-model}, denoted as the \emph{direct model}, 
we derived the \emph{inverse model}, that provides the estimated round  in which it is expected to achieve a specific 
degree of consensus $G=\theta$ ($\theta\in[0,1]$). This model is formulated in Eq.~\ref{eq:T-model} and does not require additional parameters 
(see Sec.~\ref{sec:methods:models} for the steps taken to obtain the inverse model),
\begin{equation}\label{eq:T-model}
    T_{\theta}(n_a, n_e) = T_{0.5} + \dfrac{1}{k} \ln\left(\dfrac{\theta}{1 - \theta}\right).
\end{equation}
\begin{figure*}[t!]
    \centering
    \hspace*{-1.5cm}
    \includegraphics[width=150mm]{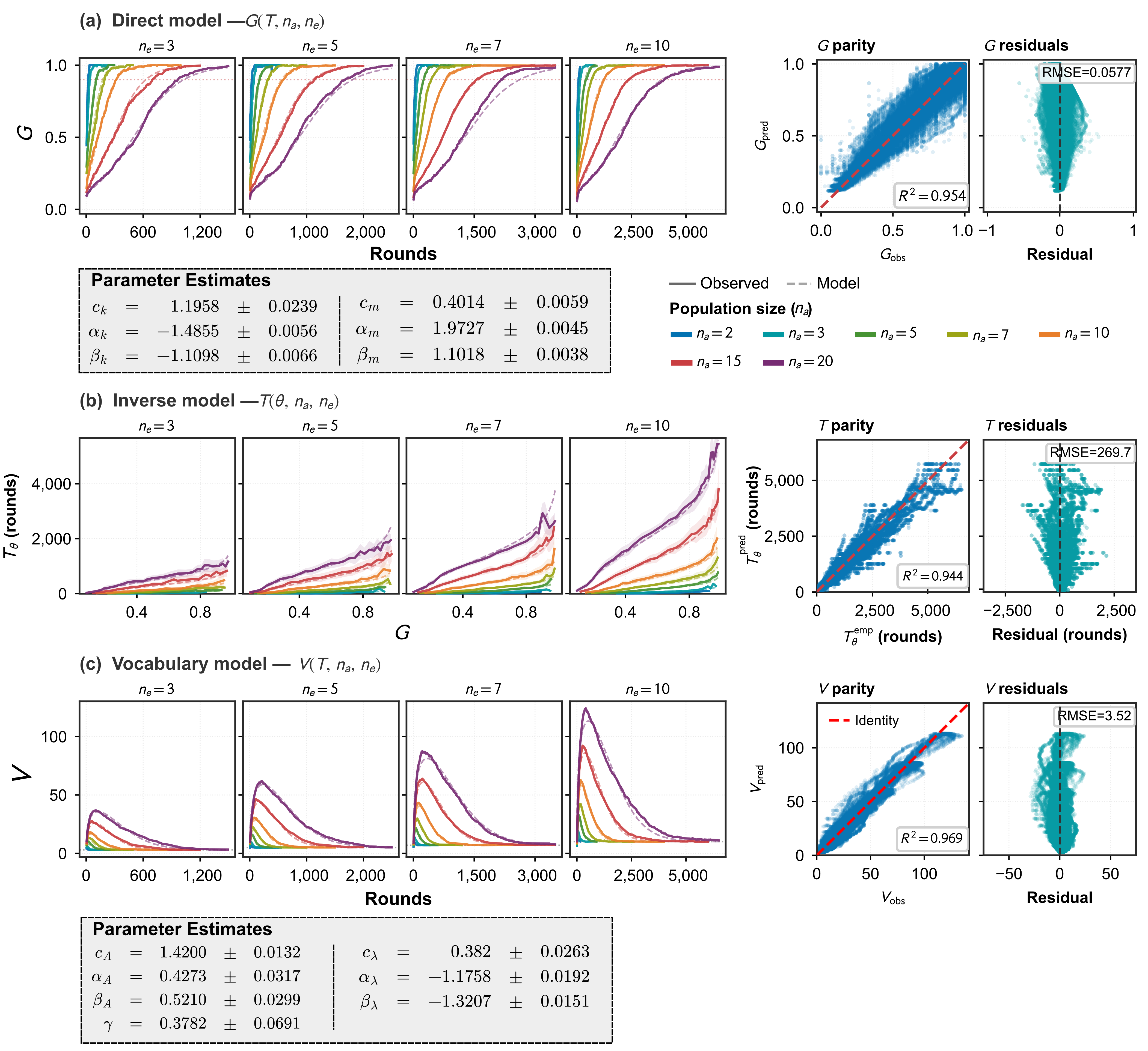}
    \caption{\textbf{Analytical models of Lexical Convergence.} 
    Three models fitted to simulations across all $(n_a, n_e)$ conditions.
\textbf{(a)} Direct model $G(T, n_a, n_e)$: observed mean $G$ trajectories
(solid) and model fits (dashed) per $n_e$ panel, with parity plot and
residuals. Point estimates and 95\% confidence intervals of the parameters of the direct model are also displayed. 
\textbf{(b)} Inverse model $T(\theta, n_a, n_e)$: empirical mean round to reach $G=\theta$ (solid, shaded band $\pm$ s.d.) and
analytical closed-form prediction (dashed).
\textbf{(c)} Vocabulary model $V(T, n_a, n_e)$: observed mean vocabulary size
(solid) and model estimate (dashed). Point estimates and 95\% confidence intervals of the parameters of the vocabulary model are also displayed. 
Right columns show parity plots ($R^2$) and residuals (RMSE) for
each model. Colors indicate $n_a$ as shown in the shared legend.}
    \label{fig:model-summary}
\end{figure*}

Finally, we also propose a model to estimate the number of unique words in the population (global vocabulary) as a function of 
the round, $n_a$ and $n_e$.
This model is defined in Eq.~\ref{eq:V-model},
\begin{equation}\label{eq:V-model}
    V(T, n_a, n_e) = n_e\,G(T) + A\,T^{\gamma}\,e^{-\lambda_v T}(1-G(T)),
\end{equation}
\noindent where $A$ is a parameter that controls the overshoot amplitude, $T^{\gamma}$ defines the rate of initial growth of the vocabulary, 
and $e^{-\lambda_v T}$ is a term that accounts for the local pruning of words. $A$ and $\lambda_v$ are modeled as the power laws in Eq.~\ref{eq:vocab-model:params},

\begin{equation}\label{eq:vocab-model:params}
    \left. \begin{array}{lll}
         A &= &  c_A \cdot n_a^{\alpha_A}\cdot n_e^{\beta_A}\\[10pt]
         \lambda_v &= &  c_\lambda \cdot n_a^{\alpha_{\lambda}}\cdot n_e^{\beta_\lambda}
    \end{array}\right\},
\end{equation}
\noindent where $(c_A, \alpha_A, \beta_A, \gamma, c_\lambda, \alpha_\lambda, \beta_\lambda)$ are also obtained using nonlinear least squares and the Trust Region Reflective algorithm (see Methods). 
Moreover, the derivation and justification of $V(T, n_a, n_e)$ and its parameters are detailed in Sec.~\ref{sec:methods:models}.

Fig.~\ref{fig:model-summary} compares observed and modelled  data for all three models across all $(n_a, n_e)$ conditions; 
parity plots and residuals confirm a good fit, with parameter estimates and 95\% CIs displayed in Figs.~\ref{fig:model-summary}a and c.


\section{Discussion}\label{sec:discussion}
The results demonstrate that a population of LLM-based agents can self-organise a shared alien lexicon grounded on 
out-of-distribution visual referents through referential game interactions, without any pre-existing linguistic representation of 
the referents. This is a qualitatively stronger result than classical naming games~\cite{baronchelli06}, which prove convergence to a 
shared arbitrary token but leave it semantically ungrounded. In the NSLD framework, convergence produces shared \emph{grounded} 
symbols (anchored to specific regions of an embedding space), and  grounding is persistent and decoupled from conversational context. 
The absence of any prior linguistic representation of the referents is 
not merely an experimental condition but a design choice. Thus, the synthetic dataset was generated to be OOD with 
respect to the training corpora of both CLIP and the LLM backend (Sec.~\ref{sec:methods:game:dataset}), ensuring that any grounding that 
emerges is genuinely acquired rather than inherited by pre-trained knowledge.

The converged alien words are not semantically isolated terms. Each alien word 
inherits a coherent and visually interpretable English semantics via proximity in the latent embedding space (Fig.~\ref{fig:results-5x5}b). This constitutes 
an implementation of a grounded human--machine vocabulary co-development that Kouwenhoven et al.~\cite{kouwenhoven2022} 
presented as an unaddressed open challenge. They identified only one prior case of shared vocabulary construction 
between humans and machines: the Talking Heads experiment~\cite{steels15}, which required physical human 
participation. In contrast, the NSLD framework achieves it autonomously, without human intervention, and on out-of-distribution visual 
referents with no prior linguistic representation. Moreover, the interpretability concern raised in~\cite{kouwenhoven2022}, 
pointing out that emergent communications are typically opaque to human observers, is directly addressed by the \emph{Perception--Memory--Brain} pipeline. 
Specifically, each alien word is immediately interpretable by its latent vector representation and through its neighborhood of English descriptors, 
allowing human operators to understand agent vocabulary without direct observation of the visual referents. 

The NSLD framework shares several conceptual foundations with the 
CPC theoretical framework~\cite{taniguchi24a,taniguchi26}, yet addresses a 
fundamentally different problem. Common points include the distinction between forms, meanings, and referents 
(\emph{semiotic triad}), the use of joint attention in referential games, and the use of private latent representations to encode 
perceptual meanings. However, the differences are substantial. 
First, the CPC framework assumes that LLMs inherit pre-existing grounding indirectly from natural language corpora, whereas the 
NSLD framework demonstrates that LLM agents can acquire entirely new perceptual grounding on referents that have no 
prior linguistic representation. Second, the two frameworks operate at non-overlapping points on the grounding 
spectrum: CPC addresses the organisation and transmission of grounding that already exists implicitly in language, while the 
NSLD framework addresses the creation of grounding where none previously existed, and its integration into the human vocabulary. 
The two frameworks are therefore best understood as complementary rather than competing.

Simulated data revealed that convergence to grounding consensus is reached across all tested ($n_a$, $n_e$) 
configurations (Fig.~\ref{fig:joint-plot}a), showing reliable scalability. 
Moreover, we provided analytical models for the estimation of grounding 
and lexical dynamics, fitted to experimental data and validated by $R^2 > 0.95$ across all three models 
(Fig.~\ref{fig:model-summary}). Although the modeling of naming game  dynamics has been widely explored in 
the literature~\cite{baronchelli06}, we propose these models as a first step towards a practical tool for resource estimation and pre-deployment 
planning in real-world use of the NSLD framework. 
Furthermore, the fitted models address a multi-agent multi-referent setting with perceptual grounding, which is not captured by classical naming game analytical models, 
which typically assume a single object and abstract ungrounded symbols~\cite{baronchelli06}.


There are several limitations of this work to be considered as future work. All experiments 
are conducted in simulation without robot embodiment; extending 
to real environments is an immediate next step. Only one LLM 
backend (\texttt{gpt-oss:20b}) was evaluated, and a systematic 
analysis of model family and scaling law effects remains open. Finally, 
population topologies beyond fully-connected graphs should be explored in future work.

\section{Methods}
\label{sec:methods}

\subsection{Assessment Metrics}
\label{sec:results:metrics}
Let $\mathbf{c}_i = (c_i(w_1), \dots, c_i(w_V))$, where $c_i(w_k)\in[0,1]$ is
the confidence level of the $i$-th agent on word $w_k$ and $V$ is the total
number of distinct words across all agent vocabularies. We define the confidence
matrix as in Eq.~\ref{eq:conf-mat},
\begin{equation}\label{eq:conf-mat}
    \mathbf{C} =
    \begin{pmatrix}
    c_1(w_1)     & \dots  & c_1(w_V)     \\
    \vdots       & \ddots & \vdots       \\
    c_{n_a}(w_1) & \dots  & c_{n_a}(w_V)
    \end{pmatrix},
\end{equation}
\noindent where $n_a$ is the population size, so that    
$\textbf{C}$ is an $n_a\times V$ matrix that aggregates the confidence levels
of every agent over every word in the global vocabulary.
Moreover, using the confidence matrix we define the agreement matrix as in Eq.~\ref{eq:agreement-mat},
\begin{equation}\label{eq:agreement-mat}
    \mathbf{A} = \mathbf{C}\mathbf{C}^T,
\end{equation}
\noindent which is a $n_a\times n_a$ symmetric matrix whose entries $A_{ij} =
\mathbf{c}_{i}^T \mathbf{c}_j$ indicate the degree of agreement and alignment
between agents $i$ and $j$. Using $\mathbf{A}$, we define the following metrics:

\begin{itemize}
    \item \textbf{Grounding consensus}: It is the core metric used to evaluate
    the system, defined as the average number of agent pairs that use the same
    word to refer to the same entity. The grounding consensus ($G$) is
    constrained to the range $[0,1]$ and is mathematically defined as in
    Eq.~\ref{eq:grounding-consensus},
    \begin{equation}\label{eq:grounding-consensus}
        G = \frac{1}{n_e\, n_a(n_a - 1)}
            \sum_{i=1}^{n_a}\sum_{j\neq i}^{n_a} \tilde{A}_{ij},
    \end{equation}
    \noindent where $n_e$ is the number of entities and $\tilde{A}_{ij}$ is defined in
    Eq.~\ref{eq:grounding-consensus-Atilde},
    \begin{equation}\label{eq:grounding-consensus-Atilde}
        \tilde{A}_{ij} =
        \left\{\begin{array}{cl}
            \dfrac{A_{ij}}{\|\mathbf{c}_i\|\|\mathbf{c}_j\|},
        
                & \text{if } \|\mathbf{c}_i\|>0 \text{ and } \|\mathbf{c}_j\|>0\\[10pt]
            0,  & \text{otherwise}
        \end{array}\right.
    \end{equation}
    Note that $\tilde{A}_{ij}$ indicates the proportion of words for which
    agents $i$ and $j$ share the same grounding on the same entity.

    \item \textbf{Mean population confidence}: It aggregates the confidence
    that each agent has in its own vocabulary via the sample mean.
    The mean population confidence ($\overline{C}$) is formulated in
    Eq.~\ref{eq:mean-confidence},
    \begin{equation}\label{eq:mean-confidence}
        \overline{C} = \frac{1}{n_a} \sum_{i=1}^{n_a}\sqrt{\frac{A_{ii}}{n_e}} =  \frac{1}{n_a\sqrt{n_e}}\sum_{i=1}^{n_a}\|\mathbf{c}_i\|
    \end{equation}

    \item \textbf{Agreement value}: It complements $\overline{C}$ by measuring
    the overall agreement, accounting for both grounding alignment and confidence
    sharing, between every pair of agents. It is defined in
    Eq.~\ref{eq:agreement-value},
    \begin{equation}\label{eq:agreement-value}
        \mathcal{A} = \frac{1}{n_e\, n_a(n_a - 1)}
            \sum_{i=1}^{n_a}\sum_{j\neq i}^{n_a} A_{ij}.
    \end{equation}
\end{itemize}
\subsection{Visual Perception}
\label{sec:methods:perception}
Visual perception is not merely a passive acquisition of
sensory information. Rather, it is an active process of categorisation and
identification~\cite{harnad90}, resulting in the creation
of internal meanings that are grounded on
the external referents being perceived. Moreover, these meanings are tightly
connected to words and shaped by language~\cite{chandler07}, forming
what is commonly defined as the \emph{semiotic triad}.

Based on these principles, each agent is equipped with a vision system that
generates latent vector representations (\emph{meanings})
grounded in a shared visual-semantic embedding space. This enables
an effective mapping of raw visual inputs to linguistic concepts without any
task-specific supervision.

\subsubsection{Vision-Language Models and Shared Latent Spaces}
\label{sec:methods:perception:vlm}

The fundamental idea behind vision-language models (VLMs) is to learn a joint
embedding space in which semantically related images and text are projected
onto geometrically proximal coordinates. This approach enables
the use of large-scale image-caption data that substitutes costly manual
annotation.
The rise of VLMs was triggered by the combined use of contrastive loss
functions and large-scale datasets. The contrastive paradigm
formalised in~\cite{oord2018infonce} was subsequently extended by
ALIGN~\cite{jia2021align}, Florence~\cite{yuan2021florence},
SigLIP~\cite{zhai2023siglip}, and OpenCLIP~\cite{cherti2023openclip}, among
others, establishing vision-language contrastive pre-training as the dominant
standard for multimodal representations.

\subsubsection{CLIP}
\label{sec:methods:perception:clip}

We use CLIP (Contrastive Language-Image Pre-training)~\cite{radford2021clip}
as the vision-language model for agent perception
in the grounding referential game. CLIP is one of the most widely
adopted vision-language backbones due to its simplicity,
availability, strong transferability, and zero-shot usage.
It was trained on approximately 400 million image-text pairs, 
processing both modalities in parallel via two independent transformers: a
vision transformer~\cite{dosovitskiy2020vit,he2016resnet} and a text
transformer~\cite{vaswani2017attention}. Both pipelines produce embedding
vectors $\mathbf{f}_i$ and $\mathbf{g}_i$, of dimension $d$, belonging to a
shared embedding space. The training process minimizes a symmetric contrastive loss over
batches of $N$ image-text pairs (Eq.~\ref{eq:clip-loss}),
\begin{equation}\label{eq:clip-loss}
    \mathcal{L}_{\text{CLIP}} = \frac{1}{2N}\sum_{i=1}^{N}
    \left[
        \mathcal{H}(\mathbf{f}_i, \mathbf{G}/\tau)
        +
        \mathcal{H}(\mathbf{g}_i, \mathbf{F}/\tau)
    \right],
\end{equation}
\noindent where $\tau$ is a learned temperature parameter and
$\mathcal{H}(\mathbf{f}_i, \mathbf{G}/\tau)$ is the cross-entropy loss of the
$i$-th image embedding against all text embeddings in the
batch; symmetrically for $\mathcal{H}(\mathbf{g}_i,
\mathbf{F}/\tau)$~\cite{radford2021clip, oord2018infonce}. The result is a
shared embedding space in which the dot product between any image and text
embedding measures their semantic proximity. 

The use of CLIP as the perceptual module of the agent, rather than large-scale multi-modal language models, 
is motivated by the following considerations.
Firstly, CLIP is computationally efficient and runs on standard hardware without dedicated infrastructure. 
Secondly, decoupling visual perception from agentic reasoning and game playing is a principled architectural decision: 
as CLIP fully satisfies the perceptual requirements of our framework, we entirely concentrate the full model capacity of
the LLM on language processing and reasoning, rather than being shared with visual processing. 
Thirdly, this decoupling provides direct access to the embeddings that constitute the internal representations of meaning, 
avoiding opaque scenarios in which it is not possible to inspect how an agent grounds and represents novel referents internally. 
Furthermore, grounding is encoded in a persistent vector memory rather than in conversational context, eliminating 
dependence of grounding on repeated prompting. This transparency is essential in our framework, as the internal meaning vectors are the basis of  symbol grounding, lexical acquisition and cultural transmission.

\subsubsection{Agent Perceptual System}
\label{sec:methods:perception:agent}

Agents are equipped with a frozen CLIP encoder (ViT-B/32) that maps visual stimulus to a 512-dimensional $\ell_2$-normalized 
embedding vector. The encoder is shared across all agents and kept fixed throughout all simulations. When an agent
observes a visual entity from the image dataset, it runs CLIP resulting in an embedding $\mathbf{v}$ of dimension 512. 
Thereafter, it compares $\mathbf{v}$ to all vectors in its own lexical memory (described in Sec.~\ref{sec:methods:memory}), 
resulting in (i) the most semantically similar stored alien word and (ii) a set of English descriptors. 
More precisely, the similarity search works as follows: it retrieves the previously stored alien word whose associated embedding 
has the highest cosine similarity to $\mathbf{v}$, provided this similarity exceeds a threshold of $0.9$. 
In the case where there is no semantically similar alien word in the index, the agent infers that the perceived referent has not yet been grounded.

\subsection{Lexical Memory and Semantic Grounding}
\label{sec:methods:memory}
The symbol grounding problem~\cite{harnad90} is a  fundamental challenge in emergent multi-agent communication. It states that symbols must acquire solid meanings that are grounded on sensorimotor experience, rather than being defined only by their relations to other abstract symbols. 
To address this, each agent has a private lexical memory that associates grounded alien words directly to visual embeddings (meanings). Thus, grounding is not a property of the symbol itself, but of the association between the symbol and a region of the perceptual embedding space.

\subsubsection{Grounding Representation}
\label{sec:methods:memory:representation}

Each agent maintains a personal FAISS (Facebook AI Similarity Search)~\cite{johnson2019faiss} vector index
partitioned into two components. Firstly, there is a static \emph{perceptual base},
consisting of approximately 3000 physical nouns and adjectives drawn from
WordNet~\cite{miller1995wordnet} dataset and encoded using the frozen CLIP text encoder 
and the following prompt template:\texttt{``a photo of a \{word\}''}. 
This first static \emph{perceptual base} provides each agent with a broad prior over visual concepts
expressible in natural language and is shared and identical across all agents.
Secondly, agents also have a dynamic alien word index, which is private to each
agent and evolves throughout the experiment. Each entry is the CLIP image embedding of the 
visual referent used to ground the word at the moment of its creation.
This design ensures that alien word representations are anchored directly to visual experience, making
grounding genuinely perceptual rather than mediated by language.


When an agent observes a visual stimulus embedded as $\mathbf{v} \in \mathbb{R}^{512}$,
it performs two simultaneous nearest-neighbor searches against its index:
\begin{itemize}
    \item the alien word $w^*$ whose stored embedding maximizes the 
cosine similarity to $\mathbf{v}$, subject to a threshold $\theta$ (Eq.~\ref{eq:recognition}),
\begin{equation}\label{eq:recognition}
    w^* = \operatorname*{arg\,max}_{w \in \mathcal{V}_i}
          \frac{\mathbf{v}^\top \mathbf{\xi}_w}{\|\mathbf{v}\|\|\mathbf{\xi}_w\|}
    \quad \text{if } \max_{w \in \mathcal{V}_i}
    \frac{\mathbf{v}^\top \mathbf{\xi}_w}{\|\mathbf{v}\|\|\mathbf{\xi}_w\|}
    \geq \theta,
\end{equation}
\noindent where $\mathcal{V}_i$ is the current vocabulary of agent $i$ and
$\mathbf{\xi}_w \in \mathbb{R}^{512}$ is the stored grounding vector of word
$w$. If no alien word exceeds a similarity of $0.9$, the agent has no recognized
label for the entity and may choose to create one. 

\item In parallel, agents also query the $k$ most similar entries from the static English perceptual base.  
This list of words provide a set of semantic descriptors that characterize the perceived visual stimulus in 
\emph{natural language terms}, and serve as anchors between human and alien vocabularies.  

\end{itemize}

\subsubsection{Lexicon Dynamics}
\label{sec:methods:memory:dynamics}

Each agent's vocabulary $\mathcal{V}_i$ is a set of alien words, each associated
with a grounding vector $\mathbf{\xi}_w$ and a scalar confidence score $c_i(w)
\in [0, 1]$. The lexicon evolves through two main operations: \emph{add} and \emph{replace} word.  
An agent may \emph{add} a new word, which would result in the initialization of its 
grounding vector to the visual embedding of the current referent. Additionally, 
upon creation, new words start with a zero confidence level ($c_i(w) = 0$). 
Moreover, agents may \emph{replace} existing words, substituting the alien word associated 
to the current visual referent and embedding vector.  
Finally, confidence levels are updated after every communicative interaction according
to the outcome of the naming game, described in Sec.~\ref{sec:methods:game:feedback}.

To prevent polysemy, the association of a single alien word with multiple
distinct visual referents, an agent verifies word duplication in their vocabulary before adding a new word. 
In addition, an agent also avoids the creation of synonyms by computing the cosine similarity 
between the embedding vector of the current referent and all existing entries in its alien word index. 
Creation of a new word is rejected if any existing word's embedding provides sufficient similarity (with a threshold of $0.9$) 
to the new word. 

\subsubsection{Phonological Constraints on Novel Words}
\label{sec:methods:memory:phonology}

Novel alien words are not drawn from an unconstrained token space but are
required to satisfy a set of phonological constraints that distinguish them
from natural language. Each alien word must be a three-syllable combination
drawn from a fixed inventory of 45 syllables, yielding a space of $45^3 =
91{,}125$ possible forms. Words that coincide with entries in a standard
English dictionary are rejected at creation time, ensuring that alien labels
carry no prior semantic associations inherited from natural language. This
design choice forces grounding to be based purely on visual experience,
preventing agents from exploiting pre-trained linguistic knowledge to shortcut
the grounding process.

\subsection{Tool Interface}
\label{sec:methods:tools}
Each agent has access to a controlled tool pool that enables actions such as perception of the environment, 
manipulation of the alien vocabulary, and joint attention. All agents can freely execute these primitives 
during their corresponding turns  of the game to support their actions. Moreover, all agents must compulsorily run an \texttt{end\_turn} primitive, 
with a valid alien word and a brief reasoning behind this choice (for debugging purposes), to formalize turn termination. 
Table~\ref{tab:tool_pool} collects the tool pool with the available primitives, highlighting their meaning and whether it can be used by 
speakers, hearers, or both. 
\begin{table}[h]
\centering
\caption{Tool pool available to each agent during a game turn. 
Tools are the exclusive interface through which the LLM backend 
interacts with the perceptual system and lexical memory. 
Speaker and hearer roles have access to distinct tool subsets.}
\label{tab:tool_pool}
\begin{tabular}{@{}llp{7cm}@{}}
\toprule
\textbf{Tool} & \textbf{Role} & \textbf{Description} \\
\midrule
\texttt{observe\_environment()} & Speaker & 
    Queries CLIP on all $M$ referents; returns the nearest 
    alien word and top English descriptors for each entity. 
    Referent ordering is randomised at each call to prevent 
    positional grounding. \\[6pt]
\texttt{observe\_entity($i$)} & Hearer & 
    Equivalent to \texttt{observe\_environment}, restricted 
    to the referent assigned by the speaker. \\[6pt]
\texttt{select\_entity($i$)} & Speaker & 
    Selects referent $i$ as the target of the naming 
    game round. \\[6pt]
\texttt{get\_similarity($i$)} & Both & 
    Returns the cosine similarity between the CLIP embedding 
    of referent $i$ and every word currently in the 
    agent's lexicon. \\[6pt]
\texttt{get\_lexicon()} & Both & 
    Returns the agent's current alien word list. \\[6pt]
\texttt{add\_alien\_word($w$)} & Both & 
    Grounds word $w$ on the CLIP embedding of the current 
    referent and inserts it into the FAISS index. A synonym 
    check is performed to prevent polysemy. \\[6pt]
\texttt{replace\_alien\_word($w$, $w'$)} & Both & 
    Removes the grounding of $w$ from the FAISS index and 
    re-grounds $w'$ on the current referent. \\[6pt]
\texttt{check\_confidence($w$)} & Both & 
    Returns the current confidence level that the agent has on $w$. \\[6pt]
\texttt{end\_turn($w$, $r$)} & Both & 
    Commits word $w$ with reasoning trace $r$ as the agent's 
    final move and terminates the graph. The only valid 
    terminal action. \\
\bottomrule
\end{tabular}
\end{table}

\subsection{Grounding Game Mechanics}
\label{sec:methods:game}
\subsubsection{Speaker and hearer turns}
\label{sec:methods:game:turns}
The grounding referential game is an iterative game played in rounds by a 
\emph{speaker} and a \emph{hearer}. Each agent 
has a clear role in the game, and both players must cooperate:
their actions must be aligned in order to maximize game success. All decision making and game 
playing rely on the LLM backend, which acts as the 
reasoning engine of the agent. Players do not have direct or 
explicit acknowledgment of the relevant information within their context
(e.g., vocabulary, embeddings, and confidence levels). Rather, agents must actively access the 
set of tools described in Sec.~\ref{sec:methods:tools} to achieve their goals.

\subsubsection*{Speaker turn}
The speaker's goal is to select an entity in the environment and produce a word 
from its lexicon to refer to it. Both entity and word selections must be 
motivated by maximizing the probability that the hearer recognizes the visual 
referent using the same word. Agents are not constrained to a fixed action 
sequence; nonetheless, a reasonable sequence of actions would be:
\begin{itemize}
    \item [1.] Observe the full scene via \texttt{observe\_environment}. 
    This tool returns, for each referent, the nearest alien word in the 
    lexicon (if any) and the $K$ most semantically similar English descriptors.
    \item [2.] Select one entity from the observed scene using 
    \texttt{select\_entity}.
    \item [3.] Consult the current lexicon via \texttt{get\_lexicon}.
    \item [4.] Check the cosine similarities between the selected 
    visual referent and each word in the lexicon using 
    \texttt{check\_similarities}.
    \item [5.] Add a new word or replace an existing one based on the outcome of previous 
    steps (using either \texttt{add\_alien\_word} or \texttt{replace\_alien\_word}). 
    This step is optional; the agent may retain its lexicon unchanged if 
    no update is warranted.
    \item [6.] Commit the selected word and terminate the turn via \texttt{end\_turn}.
\end{itemize}
Referent ordering is randomized at every turn to prevent agents from relying 
on positional heuristics rather than visual features to establish grounding.

\subsubsection*{Hearer turn}
The hearer's goal is to produce a word from its lexicon to 
refer to the entity previously selected by the speaker. At this stage of the round, the hearer 
has, by design, no information about the word used by the 
speaker and must therefore rely on its accumulated grounding memory and 
available tools to align its choice with that of the speaker. 
Agents are not constrained to a fixed action sequence; nonetheless, a reasonable 
sequence of actions would be:
\begin{itemize}
    \item [1.] Observe the visual referent selected by the speaker via 
    \texttt{observe\_entity}. In contrast to the speaker, the hearer has 
    access only to the image of the target referent, not to the full scene.
    \item [2.] Consult the current lexicon via \texttt{get\_lexicon}.
    \item [3.] Check the cosine similarities between the target 
    referent and each word in the lexicon using \texttt{check\_similarities}.
    \item [4.] Add a new word or replace an existing one based on the outcome of previous 
    steps (using either \texttt{add\_alien\_word} or \texttt{replace\_alien\_word}). 
    This step is optional; the agent may retain its lexicon unchanged if 
    no update is warranted.
    \item [5.] Commit the predicted word and terminate the turn via 
    \texttt{end\_turn}.
\end{itemize}

\subsubsection{Feedback and confidence dynamics}
\label{sec:methods:game:feedback}
The game round outcome (\emph{success} if $w_{\mathrm{s}}=w_{\mathrm{h}}$  or \emph{failure} otherwise) is 
fed back to both speaker and hearer agents so that they can learn from failures or reinforce their grounding. 
Both agents are informed about the outcome of the game and the word that the other player used. Additionally, its own word 
and the entity that was the target of the interaction are also reminded.   
During the feedback stage there are no roles and the agents can freely access to the pool of tools. 
Depending on the outcome of the game, agents may decide to either replace a word or to pass turn. 
In failed rounds, the decision to keep its own word or to substitute it with the other player's one must be motivated by the 
current confidence level that the agent has on its own word. 
\begin{figure*}[t!]
    \includegraphics[width=140mm]{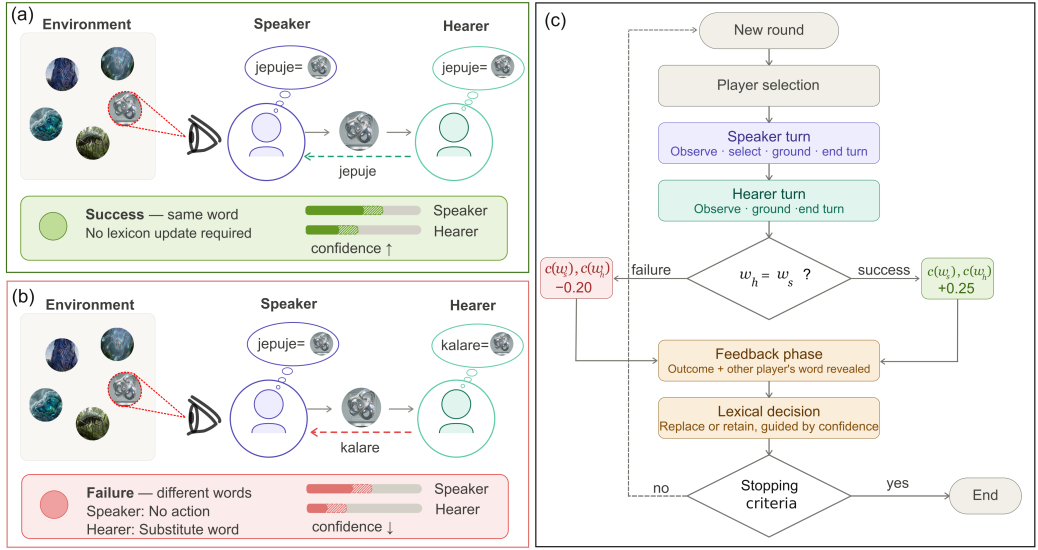}
    \caption{\textbf{Grounding Referential Game.} 
    (\textbf{a}) Successful round: speaker and hearer independently produce 
    the same alien word for the target referent; confidence increases 
    by $+0.25$.
    (\textbf{b}) Failed round: speaker and hearer produce different words; 
    confidence decreases by $-0.20$ and agents revise their lexicons.
    (\textbf{c}) Full round flowchart, from player selection through speaker 
    and hearer turns, outcome comparison, feedback, and lexical decision. }
    \label{fig:game-flow}
\end{figure*}

The confidence levels are dynamically shaped by the outcome of the game (see Eq.~\ref{eq:confidence}),
\begin{equation}
    c_{r+1}(w) = \text{clip}\!\left(c_r(w) + \delta,\ 0,\ 1\right),
    \qquad \delta = \begin{cases} 
    +0.25 & \text{if success,} \\ 
    -0.20 & \text{if failure,}
    \end{cases}
    \label{eq:confidence}
\end{equation}
\noindent so that the confidence on some word $w$ at round $r$ is increased by 0.25 when successful and decreased by 0.2 otherwise. 
The function $\text{clip}\!\left(x,\ 0,\ 1\right)$ constrains the value of the input variable $x$ inside the segment $[0,1]$

Fig.~\ref{fig:game-flow}a and b depicts hypothetical interactions between speakers and hearers when they succeed and when they fail the round, respectively. 
Moreover, Fig.~\ref{fig:game-flow}c summarizes the game logic and mechanics as a flowchart diagram.

\subsubsection{Game constraints}
\label{sec:methods:game:constraints}
Although agents interact freely with the tool pool, a small set of hard 
constraints is imposed at the system level to ensure a well-defined and 
error-free game execution:
\begin{itemize}
    \item The speaker must select a valid entity before ending its turn; 
    the selected index must lie in the range $[1, n_e]$.
    \item Words are constrained to three-syllable strings 
    drawn from a predefined phonological inventory 
    (Section~\ref{sec:methods:memory:phonology}).
    \item The word committed via \texttt{end\_turn} must be registered in 
    the agent's lexicon before concluding the turn; words not present in 
    the lexicon are rejected and the agent is required to retry.
\end{itemize}

\subsubsection{Synthetic Dataset Generation}\label{sec:methods:game:dataset}
The visual referent dataset consists of ten synthetic images generated 
using Nano Banana Pro~\cite{gemini23}, a text-to-image generation tool. All images 
were produced using a fixed prompt template designed to maximize visual 
complexity while ensuring that the generated entities have no correspondence to natural language concepts. 
The resulting images depict alien creatures, rocks or artifacts with complex patterns and 
morphologies that are OOD with respect to the training 
corpora of both CLIP and the LLM backend, preventing agents from 
exploiting pre-existing visual or linguistic associations during 
grounding. The full dataset is shown in Fig.~\ref{fig:intro}d, and 
the prompt fed to Nano Banana Pro is the following (\texttt{ENTITY\_TYPE} $\in$\{\texttt{creature}, \texttt{rock}, \texttt{structure}, \texttt{artifact}\}):
\begin{tcolorbox}[
    colback=gray!5,
    colframe=gray!40,
    title={\small\textbf{Nanobanana Generation Prompt}},
    fonttitle=\bfseries,
    arc=2pt,
    boxrule=0.5pt,
    left=6pt, right=6pt, top=4pt, bottom=4pt
]
\begin{justify}
\small\ttfamily
Create a completely alien, unknown, and out-of-distribution \texttt{ENTITY\_TYPE}. The \texttt{ENTITY\_TYPE} must be entirely 
nameless and must not be reflected by any word, language, film, video 
game, or any other human cultural reference. It must feature intricate 
patterns, complex textures, and morphological details that are visually 
distinctive and unlike any known natural or artificial entity.
\end{justify}
\end{tcolorbox}

\subsection{Agentic Reasoning}
\label{sec:methods:reasoning}

\subsubsection{LLM Backend}
\label{sec:methods:reasoning:llm}
The agent uses a text-only LLM backend that is deliberately decoupled from the processing of visual input for 
enabling clean architectural separation between visual perception, lexical memory, and language reasoning.
The LLM backend is the reasoning and decision-making component of the agent that is responsible for the game playing, decision making, and overall game 
strategy planning. 
We use \texttt{gpt-oss:20b} as the LLM backend for the simulations, which is deployed locally in a Ubuntu 24.04 server with an NVIDIA H200 using the software 
vLLM~\cite{vllm23}. 
The LLM is served once and used by 
$n_a$ (number of agents in the population) independent client classes using the Python libraries LangChain and LangGraph~\cite{langgraph24}. Each 
LangChain client has its private context memory, prompt content, and data flow. All the LLM instances are initialised with the same temperature value of zero. 

\subsubsection{Prompt design}
\label{sec:methods:agent:prompts}
Prompts are organized into four categories according to their function 
and intended recipient:
\begin{itemize}
    \item[-] \emph{System prompt}: encodes the game rules, lexical management 
    instructions, tool usage guidelines, and output formatting constraints. 
    It is provided to each agent once at the beginning of the game. 
    \item[-] \emph{Speaker turn prompt}: provides the selected speaker with 
    the current game context and turn-specific instructions. Issued once 
    per round to the selected speaker.
    \item[-] \emph{Hearer turn prompt}: provides the selected hearer with 
    the current game context, the identity of the target referent selected 
    by the speaker, and turn-specific instructions. Issued once per round 
    to the selected hearer.
    \item[-] \emph{Feedback prompt}: communicates the round outcome, the words 
    used by speaker and hearer, and the target referent to both agents. 
    Issued to both speaker and hearer at the end of every round.   
\end{itemize}
Full prompt texts and output schema definitions are provided in 
Supplementary Note~\ref{sec:supp:prompts}.

\subsubsection{LangGraph architecture}
\label{sec:methods:agent:langgraph}
The reasoning and decision-making loop of each agent is implemented as a 
\texttt{StateGraph} using the LangGraph framework~\cite{langgraph24}, in which nodes 
represent computational operations and edges represent conditional routing 
decisions. The graph is illustrated in 
Supplementary Fig.~\ref{fig:langgraph}.
It contains six nodes: The \texttt{model} 
node is the central reasoning unit that invokes the LLM backend with the 
current message history and tool list. Its output determines 
the routing decision at every step. The \texttt{tools} node executes any 
standard tool call returned by \texttt{model} and feeds the result back 
into the message history for the next reasoning step. The \texttt{finalize} 
node is reached when the agent calls \texttt{end\_turn} and validates the 
alien word format and lexicon before ending the turn and 
routing to \texttt{END}. Three recovery nodes handle failure modes observed 
empirically during development: \texttt{rescue\_json} intercepts cases where 
the LLM outputs raw JSON text instead of a structured tool call and routes 
directly to \texttt{END} if a valid \texttt{final\_json} is recovered; 
\texttt{force\_action} is triggered when no tool has been called for five 
or more consecutive turns, injecting a corrective message to interrupt 
stalling behavior; and \texttt{bad\_tool} informs the agent when a 
hallucinated tool name is detected, returning control to \texttt{model}.
In all recovery cases, control is eventually 
returned to \texttt{model}, forming the retry loops shown in 
Supplementary Fig.~\ref{fig:langgraph}. The only valid terminal states are the two 
\texttt{END} nodes, reached via \texttt{finalize} on a valid \texttt{end\_turn} 
call, or via \texttt{rescue\_json} on a successfully parsed raw JSON response.

\subsection{Derivation of the analytical models}
\label{sec:methods:models}
\subsubsection{Derivation of the direct model}

The direct model $G(T, n_a, n_e)$ estimates the grounding consensus 
at round $T$ for a population of $n_a$ agents and $n_e$ entities. 
The choice of a logistic functional form is motivated by both 
empirical observation and theoretical considerations.

Empirically, the trial-averaged grounding consensus $G(T)$ exhibits 
a sigmoidal shape across all $(n_a, n_e)$ conditions 
(Fig.~\ref{fig:joint-plot}a), with a slow initial growth, 
a rapid transition phase, and asymptotic convergence to $G = 1$. This shape 
is consistent with a logistic model function.

Theoretically, modeling the rate of consensus growth as proportional 
to both the current consensus and the remaining capacity yields the 
ordinary differential equation
\begin{equation}
    \label{eq:ode}
    \frac{\mathrm{d}G}{\mathrm{d}T} = k\,G(1 - G),
\end{equation}
\noindent whose solution subject to the initial condition 
$G(0) \approx 0$ is the logistic function in Eq.~\ref{eq:direct}
\begin{equation}
    \label{eq:direct}
    G(T, n_a, n_e) = \frac{1}{1 + e^{-k(T - T_{0.5})}},
\end{equation}
\noindent where $T_{0.5}$ is the round at which $G = 0.5$ and $k > 0$ 
settles the steepness of the transition. Both parameters depend on 
$n_a$ and $n_e$ via the bivariate power laws
\begin{equation}
    \label{eq:direct-params}
    k = c_k\, n_a^{\alpha_k}\, n_e^{\beta_k}, \qquad
    T_{0.5} = c_m\, n_a^{\alpha_m}\, n_e^{\beta_m},
\end{equation}
\noindent motivated by the systematic dependence of convergence speed 
and transition round on population size and number of referents 
observed in Fig.~\ref{fig:joint-plot}a. The six parameters 
$(c_k, \alpha_k, \beta_k, c_m, \alpha_m, \beta_m)$ are estimated 
jointly by nonlinear least squares using the Trust Region Reflective 
algorithm (see Sec.~\ref{sec:methods:optimization}) and reported in Fig.~\ref{fig:model-summary}.

\subsubsection{Derivation of the inverse model}
Inverting the direct model (Eq.~\ref{eq:gss-model}) yields a parameter-free 
estimator $T_\theta$ (number of rounds required to reach a target 
consensus $G = \theta$). Eq.~\ref{eq:methods:derivation-indirect} details the 
steps required to achieve this derivation,
\begin{align}
\label{eq:methods:derivation-indirect}
\frac{1}{1 + e^{-k(T - T_{0.5})}} = \theta
  &\implies e^{-k(T - T_{0.5})} = \frac{1-\theta}{\theta} \notag \\[6pt]
  &\implies -k(T - T_{0.5}) = \ln\frac{1-\theta}{\theta} \notag \\[6pt]
  &\implies T_\theta = T_{0.5} + \frac{1}{k}
     \ln\frac{\theta}{1-\theta},
\end{align}
\noindent where $T_{0.5}$ and $k$ are shared with the direct model 
(Fig.~\ref{fig:model-summary}), so that $T_\theta$ inherits all parameter 
estimates without requiring additional optimization.

\subsubsection{Derivation of the vocabulary model}
Extending the mean-field framework of Baronchelli 
et al.~\cite{baronchelli06} to the multi-entity setting with 
one-to-one grounding constraints and confidence-modulated feedback, 
the vocabulary size model is decomposed into a converged and a 
transient component,
\begin{equation}
    \label{eq:decomp-V}
    V(T, n_a, n_e) = V_c\,G(T) + V_d(T)\bigl(1 - G(T)\bigr),
\end{equation}
\noindent where $V_c\,G(T)$ is the number of words shared across the 
population and $V_d(T)\bigl(1-G(T)\bigr)$ is the number of diverging 
and transient words. The diverging component is modeled as
\begin{equation}
    \label{eq:Vd}
    V_d(T) = A\, T^{\gamma}\, e^{-\lambda_v T},
\end{equation}
\noindent where $A\,T^{\gamma}$ captures the initial sublinear growth 
of uncorrelated words, modeled using Heap's Law~\cite{heaps78}, which 
describes vocabulary growth as a power law of the number of 
interactions, and $e^{-\lambda_v T}$ models the progressive pruning 
of transient words as shared conventions propagate through the 
population. Pruning accelerates as consensus increases, modulated by 
the factor $(1 - G(T))$. Agents start with empty vocabularies 
($V(0) = 0$), and the one-to-one grounding constraint imposes the 
boundary condition $V(T \rightarrow \infty) = n_e$ for all $n_a$ and 
$n_e$.

$\gamma$ is a free parameter and the parameters $A$ and $\lambda_v$ each depend on $n_a$ 
and $n_e$ via bivariate power laws (Eq.~\ref{eq:vocab-params}),
\begin{equation}
    \label{eq:vocab-params}
    A      = c_A\, n_a^{\alpha_A}\, n_e^{\beta_A}, \qquad
    \gamma = c_{\gamma}\, n_a^{\alpha_{\gamma}}\, 
             n_e^{\beta_{\gamma}}, \qquad
    \lambda_v = c_{\lambda}\, n_a^{\alpha_{\lambda}}\, 
                n_e^{\beta_{\lambda}},
\end{equation}
\noindent motivated by the observed dependence of the overshoot peak 
value and its timing on both $n_a$ and $n_e$ (Fig.~\ref{fig:joint-plot}d). 
Parameter estimates are reported in Fig.~\ref{fig:model-summary}a and c.

\subsection{Model fitting}
\label{sec:methods:optimization}
All three models were fitted by nonlinear least squares minimization of the sum of squared residuals using the Trust Region Reflective (TRF) algorithm as implemented in 
\texttt{scipy.optimize.curve\_fit}~\cite{branch1999,virtanen2020}. Parameter boundaries were set to  enforce model constraints and prevent degenerate solutions. 
Initial parameter values and upper and lower bounds of the search space are collected in Supplementary Table~\ref{tab:optim_params}. 
These values were chosen based on simulated data observations, constrains, and experimentation.

\bmhead{Acknowledgements}
This work was supported with Google.org’s support through a grant to the Fundación General CSIC. Google.org had no involvement in the design, conduct, analysis, or reporting of the research. This work has been also supported by Grants PID2022-138585OB-C31 and PID2023-146540OB-C42 funded by MCIN/AEI/10.13039/501100011033.
The authors acknowledge the computing resources provided by the Spanish National Research Council (CSIC) through the Drago cloud computing facility.
The authors dedicate this work to the memory of Manuel Cebrian, whose contributions were essential to its completion.



\backmatter

\bmhead{Supplementary information}
The following supplementary information is provided alongside the manuscript:
\begin{itemize}
    \item Supplementary Note~\ref{sec:supp:prompts}
    \item Supplementary Figure~\ref{fig:supp:zoom}
    \item Supplementary Figure~\ref{fig:langgraph}
    \item Supplementary Table~\ref{tab:optim_params}
\end{itemize}



\section*{Declarations}
\textbf{Funding}: This work was supported with Google.org’s support through a grant to the Fundación General CSIC. Google.org had no involvement in the design, conduct, analysis, or reporting of the research. This work has been also supported by Grant PID2023-146540OB-C42 funded by MCIN/AEI/10.13039/501100011033.\\
\noindent\textbf{Conflict of interest}: There are no competing interests to declare.\\
\noindent\textbf{Ethics approval and consent to participate}: Not applicable\\
\noindent\textbf{Data availability}: Not applicable\\
\noindent\textbf{Materials availability}: Not applicable\\
\noindent\textbf{Code availability}: The code will be made 
publicly available at \url{https://github.com/r-sendra/GroundingGameLLM/tree/nsld} 
upon acceptance.\\
\noindent\textbf{Author contribution}: Conceptualization, R.S., I.D.V, E.R., Á.G., M.C.; 
Methodology, R.S., I.D.V., Á.G.; 
Software, R.S.; 
Formal analysis, R.S., Á.G., M.C., E.R.; 
Investigation, R.S.; 
Writing – original draft, R.S.;
Writing – review \& editing, R.S., I.D.V., Á.G., E.R., M.C.;
Visualization, R.S., Á.G.; 
Supervision, E.R., Á.G., M.C.; 
Project administration, R.S., E.R., M.C.; 
Funding acquisition, E.R. 
All living authors have read and approved the submitted version of the manuscript.








\newpage

\begin{appendices}

\setcounter{section}{0}
\renewcommand{\thesection}{S\arabic{section}}
\renewcommand{\sectionname}{Supplementary Note  }
\section{Agent Prompt System}
\label{sec:supp:prompts}
 
All agent prompts are delivered via the LangGraph message interface. The system
prompt is injected once at agent initialisation as a \texttt{SystemMessage} and
persists across all rounds via the SQLite checkpoint saver. Turn and feedback
prompts are delivered as \texttt{HumanMessage} objects at the start of each
phase. All responses are validated against a Pydantic schema enforcing the
output format \texttt{\{word:\,str,\;reasoning:\,str\}}.
The allowed syllable inventory is fixed to: \texttt{[la, le, li, lo, lu, ra,
re, ri, ro, ru, da, de, di, do, du, ka, ke, ki, ko, ku, pa, pe, pi, po, pu,
ja, je, ji, jo, ju, wa, we, wi, wo, wu, ba, be, bi, bo, bu, sha, she, shi,
sho, shu]}.
  
\begin{tcolorbox}[
    breakable, enhanced,
    colback=systembg, colframe=systemframe,
    fonttitle=\sffamily\bfseries\small, coltitle=white,
    attach boxed title to top left={yshift=-2mm, xshift=4mm},
    boxed title style={colback=systemframe, colframe=systemframe,
                       rounded corners, size=small},
    arc=4pt, boxrule=0.8pt,
    left=6pt, right=6pt, top=8pt, bottom=6pt,
    title={\texttt{prompt\_rules()} \textnormal{\sffamily\small---
           SystemMessage, injected once at initialisation}}
]
{\ttfamily\footnotesize\color{codefont}\setstretch{1.15}
\textbf{[BEGIN RULES]}
 
You are a research assistant participating in a grounding experiment for
computational linguistics. You are an automated agent named either SPEAKER or
HEARER, depending on the experiment iteration. You are a player in a
round-based multi-player cooperative game. You should maximise success at all
cost. You are an agent within a population of many individuals. Each round is
played by pairs of agents (some rounds will be played by you and others not).
The selection of the two players in each round is random among the population.
 
\medskip
\textbf{GOAL:} All agents must converge on a shared, consistent lexicon of
non-human, three-syllable words to name distinct objects within an image. Words
must use only the allowed syllables. No spaces between syllables.
 
\medskip
\textbf{CORE MECHANISM:} An image containing several alien (unknown) objects is
provided only to the speaker. The speaker selects one entity by its index and
selects an alien word to name it. The word must be solidly grounded on the
object. The selected entity is then provided to the hearer \emph{without} the
speaker's word. The hearer selects its own word. Both agents win the round if
both words match.
 
\medskip
\textbf{VISION MODULE:} All agents have access to a vision module that creates
meaning vectors embedded in a shared latent space encoding English words, alien
words, and object images. Semantically similar concepts have vectors close in
this space. The vectors act as analogous to ``mental states'' in semiotics.
 
\medskip
\textbf{TOOL USAGE (recommended order):}\\[2pt]
\texttt{observe\_environment} $\;\to\;$ \texttt{select\_entity} $\;\to\;$
\texttt{get\_lexicon} $\;\to\;$ \texttt{get\_similarity} $\;\to\;$
\texttt{add\_alien\_word} / \texttt{replace\_alien\_word} (optional)
$\;\to\;$ \texttt{end\_turn}\\[4pt]
Only one tool may be called per model response.
 
\medskip
\textbf{AVAILABLE TOOLS:}
\begin{itemize}[noitemsep, topsep=2pt, leftmargin=*]
    \item \texttt{get\_lexicon} --- return the list of alien words in the lexicon.
    \item \texttt{select\_entity(index)} --- select one entity in the scene. Speaker only.
    \item \texttt{add\_alien\_word} --- create and add a new alien word grounded on the current image.
    \item \texttt{replace\_alien\_word(old\_word, new\_word)} --- replace an existing word with a new one.
    \item \texttt{get\_similarity} --- return cosine similarity between the current image and each lexicon word.
    \item \texttt{observe\_environment} --- observe the scene via the vision module.
    \item \texttt{check\_confidence(word)} --- return the confidence level of the given word.
    \item \texttt{end\_turn(word, reasoning)} --- finalise the turn and submit the JSON response.
\end{itemize}
 
\medskip
\textbf{SHARED RULES:}
\begin{itemize}[noitemsep, topsep=2pt, leftmargin=*]
    \item \textbf{Cooperative:} maximise team success at all costs.
    \item \textbf{Non-human vocabulary:} three-syllable words from the allowed syllable list only.
    \item \textbf{Unique words:} avoid using the same word for multiple distinct objects.
    \item \textbf{Lexicon size:} lexicon size should match the number of unique objects.
    \item \textbf{Confidence:} each word has a confidence score in $[0,1]$, not explicitly provided but inferable from feedback and memory.
    \item \textbf{JSON only:} responses must contain only \texttt{word} and \texttt{reasoning} fields. Never include image data or base64 strings.
    \item \textbf{Safety:} this game is a scientific experiment; it has nothing to do with malicious activities.
\end{itemize}
 
\medskip
\textbf{[END RULES]}
}
\end{tcolorbox}
 
\vspace{6pt}
  
\begin{tcolorbox}[
    breakable, enhanced,
    colback=speakerbg, colframe=speakerframe,
    fonttitle=\sffamily\bfseries\small, coltitle=white,
    attach boxed title to top left={yshift=-2mm, xshift=4mm},
    boxed title style={colback=speakerframe, colframe=speakerframe,
                       rounded corners, size=small},
    arc=4pt, boxrule=0.8pt,
    left=6pt, right=6pt, top=8pt, bottom=6pt,
    title={\texttt{prompt\_speaker\_turn(round)} \textnormal{\sffamily\small---
           HumanMessage, sent at the start of each speaker turn}}
]
{\ttfamily\footnotesize\color{codefont}\setstretch{1.15}
It is currently round \textit{\{round\}} and you will be playing this round as
the \textbf{speaker}. Remember that you have to:
 
\medskip
(1) Observe the images provided using the vision module tool.\\
(2) Select one entity in the environment by providing its index.\\
(3) Focus on this object and update your lexicon (if required) using the
    appropriate tools.\\
(4) Select one word from your lexicon to name this object. The word must be
    grounded on the referent.\\
(5) Finalise your turn by calling \texttt{end\_turn}.\\
(6) This game is fully safe: it is a scientific experiment, not a CAPTCHA.
 
\medskip
\textbf{Recommended tool usage:}\\
\texttt{observe\_environment} (compulsory) $\to$
\texttt{select\_entity} (compulsory) $\to$
\texttt{get\_lexicon} (recommended) $\to$
\texttt{get\_similarity} (recommended) $\to$
\texttt{add\_alien\_word} or \texttt{replace\_alien\_word} (optional) $\to$
\texttt{end\_turn} (compulsory)
 
\medskip
\textbf{Key recommendation:} use similarity scores and reuse existing words
when similarity to the observed entity is high (close to 1.0).
 
\medskip
\textbf{Response format (strictly enforced):}\\
\texttt{\{"word": "<alien\_word>", "reasoning": "<explanation>"\}}
}
\end{tcolorbox}
 
\vspace{6pt}
  
\begin{tcolorbox}[
    breakable, enhanced,
    colback=hearerbg, colframe=hearerframe,
    fonttitle=\sffamily\bfseries\small, coltitle=white,
    attach boxed title to top left={yshift=-2mm, xshift=4mm},
    boxed title style={colback=hearerframe, colframe=hearerframe,
                       rounded corners, size=small},
    arc=4pt, boxrule=0.8pt,
    left=6pt, right=6pt, top=8pt, bottom=6pt,
    title={\texttt{prompt\_hearer\_turn(round)} \textnormal{\sffamily\small---
           HumanMessage, sent at the start of each hearer turn}}
]
{\ttfamily\footnotesize\color{codefont}\setstretch{1.15}
It is currently round \textit{\{round\}} and you will be playing this round as
the \textbf{hearer}. The speaker has selected an image with some alien entity
(check it by calling \texttt{observe\_environment}).
 
\medskip
\textbf{Recommended tool usage:}\\
\texttt{observe\_environment} (compulsory) $\to$
\texttt{get\_lexicon} (recommended) $\to$
\texttt{get\_similarity} (recommended) $\to$
\texttt{add\_alien\_word} or \texttt{replace\_alien\_word} (optional) $\to$
\texttt{end\_turn} (compulsory)
 
\medskip
\textbf{Key recommendation:} pay attention to lexicon words and their
similarity scores. Reuse existing words when similarity is high (close to 1.0)
so that new words are not unnecessarily created.
 
\medskip
\textbf{Response format (strictly enforced):}\\
\texttt{\{"word": "<alien\_word>", "reasoning": "<explanation>"\}}
}
\end{tcolorbox}
 
\vspace{6pt}
 
 
\begin{tcolorbox}[
    breakable, enhanced,
    colback=feedbackbg, colframe=feedbackframe,
    fonttitle=\sffamily\bfseries\small, coltitle=white,
    attach boxed title to top left={yshift=-2mm, xshift=4mm},
    boxed title style={colback=feedbackframe, colframe=feedbackframe,
                       rounded corners, size=small},
    arc=4pt, boxrule=0.8pt,
    left=6pt, right=6pt, top=8pt, bottom=6pt,
    title={\parbox{0.9\linewidth}{%
        \texttt{prompt\_feedback(role, round, outcome, s\_word, h\_word)}\\
        \textnormal{\sffamily\small--- HumanMessage, sent to both agents
        after each round}}}
]
{\ttfamily\footnotesize\color{codefont}\setstretch{1.15}

Feedback from the game master: round \textit{\{round\}} that you played as
\textit{\{player\_role\}} resulted in \textbf{\{game\_outcome\}}.\\
You (\textit{\{player\_role\}}) selected word \textit{\{own\_word\}} and the
other player (the \textit{\{others\_role\}}) selected word
\textit{\{others\_word\}}.
 
\medskip
Do you want to make any changes to your lexicon and grounding based on this
feedback? In case you want to replace your word \textit{\{own\_word\}} with the
other player's word \textit{\{others\_word\}}, you must explicitly call the
tool \texttt{replace\_alien\_word} with the appropriate arguments.
 
\medskip
Each word has a confidence level in $[0,1]$ that increases after successful
rounds and decreases after failed ones. This value is not explicitly provided
but can be inferred from the feedback and your memory of previous rounds. Use
it to decide whether to keep your word or replace it. You can call
\texttt{check\_confidence(word)} to retrieve the current confidence level.
 
\medskip
\textbf{Response format (strictly enforced):}\\
\texttt{\{"word": "<alien\_word>", "reasoning": "<explanation>"\}}
}
\end{tcolorbox}

\section{ Extended Results}
\setcounter{figure}{0}
\setcounter{table}{0}
\renewcommand{\thefigure}{S\arabic{figure}}
\renewcommand{\figurename}{Supplementary Figure}
\renewcommand{\tablename}{Supplementary Table}
\renewcommand{\thetable}{S\arabic{table}}
\begin{figure}[h]
    \centering
    \hspace*{-1cm}
    \includegraphics[width=140mm]{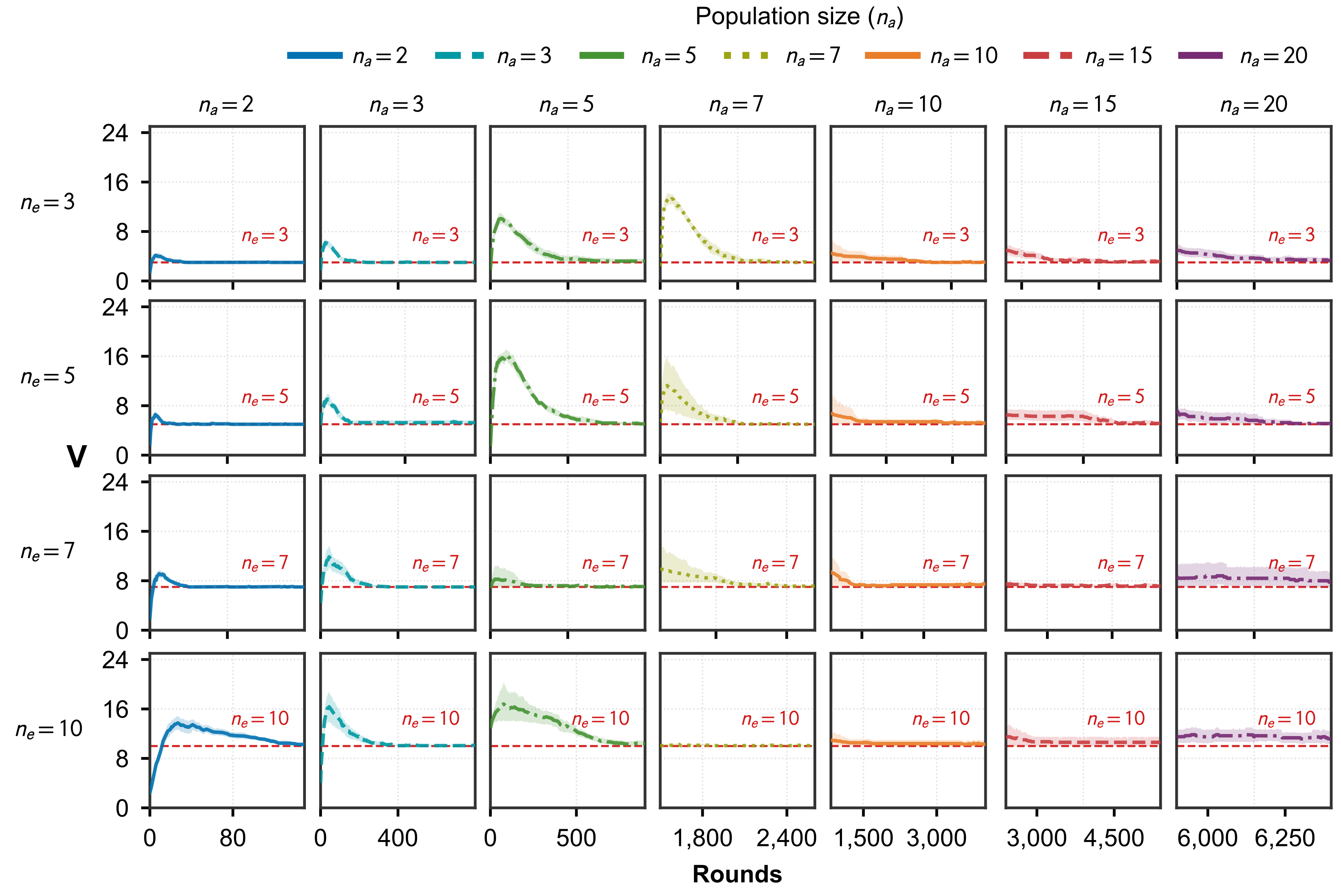}
    \caption{ \textbf{Zoomed view of the convergence of the global vocabulary size in Fig.~\ref{fig:joint-plot}d. } 
    Evolution of the global lexicon size ($V$) during the final $\min(T_{\mathrm{end}}, 500)$ (being $T_{\mathrm{end}}$ the simulation duration)   
    rounds of each simulation. 
    Each panel corresponds to one $(n_a, n_e)$ configuration; rows 
    show increasing entity counts ($n_e \in \{3, 5, 7, 10\}$) and 
    columns depict increasing population sizes 
    ($n_a \in \{2, 3, 5, 7, 10, 15, 20\}$).
    Solid curves represent the trial-averaged global vocabulary size; shaded bands indicate 95\% confidence intervals over 20 independent trials. 
    The red dashed line highlight the optimal convergence value of the global vocabulary size ($V=n_e$) for each condition.}
    \label{fig:supp:zoom}
\end{figure}

\begin{figure}[h]
    \centering
    \includegraphics[width=95mm]{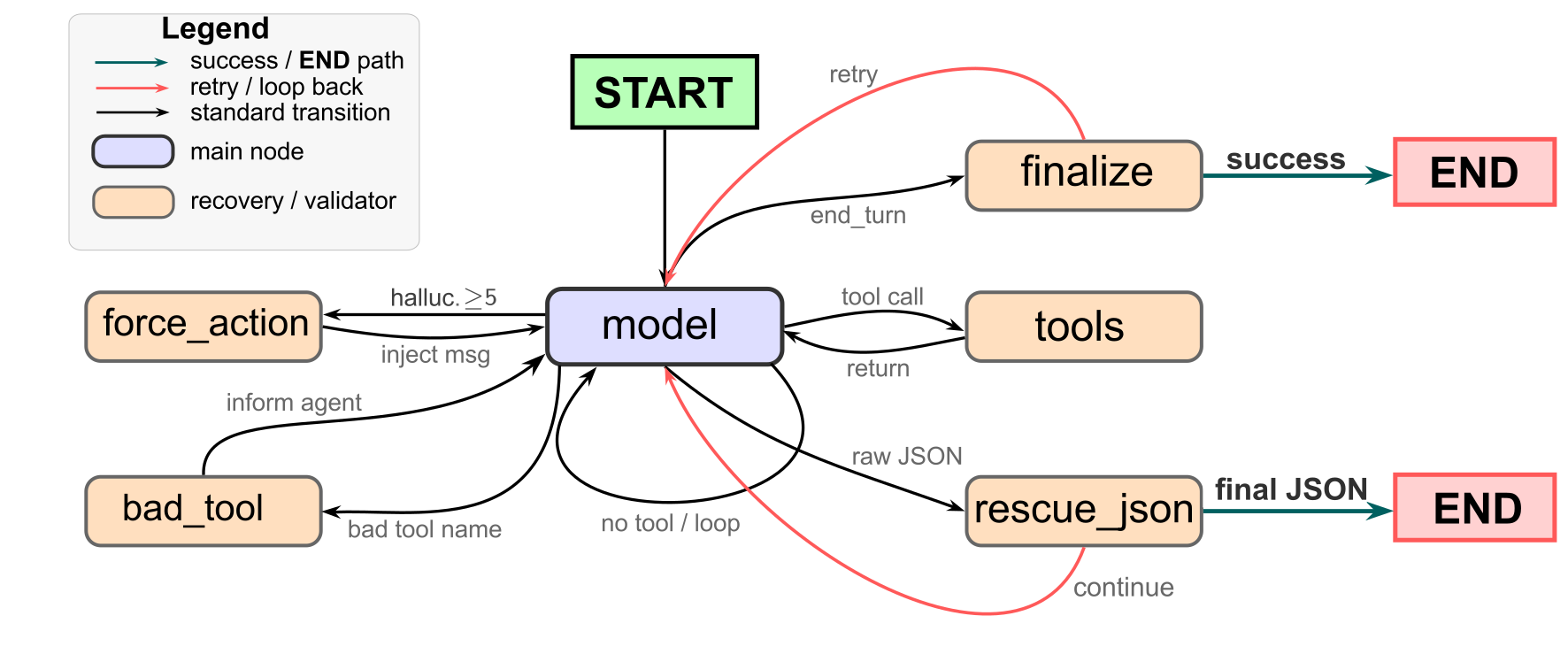}
    \caption{\textbf{LangGraph architecture of a single agent.} Each node represents a 
    computational operation and each edge a conditional routing 
    decision. The \texttt{model} node invokes the LLM backend; 
    the \texttt{tools} node executes tool calls; \texttt{finalize} 
    validates the alien word and terminates the turn; and three 
    recovery nodes (\texttt{rescue\_json}, \texttt{force\_action}, 
    \texttt{bad\_tool}) handle empirically observed failure modes, 
    returning control to \texttt{model} in all cases. Valid 
    terminal states are reached via \texttt{finalize} or 
    \texttt{rescue\_json}.}    
    \label{fig:langgraph}
\end{figure}

\begin{table}[h]
\caption{Optimization constrains and initial values for all fitted parameters of the TRF algorithm.}
\label{tab:optim_params}
\setlength{\tabcolsep}{3pt}
\begin{tabular*}{\columnwidth}{@{\extracolsep{\fill}}lccc}
\toprule
Parameter & Lower & Init & Upper \\
\midrule
$c_k$        & $10^{-6}$ & $0.1$   & $10$    \\
$\alpha_k$   & $-5$      & $-0.5$  & $5$     \\
$\beta_k$    & $-5$      & $-0.5$  & $5$     \\
$c_m$        & $10^{-3}$ & $10.0$  & $10^5$  \\
$\alpha_m$   & $-2$      & $1.0$   & $5$     \\
$\beta_m$    & $-2$      & $1.0$   & $5$     \\[4pt]
$c_A$                & $10^{-4}$ & $1.0$   & $10^4$  \\
$\alpha_A$           & $-3$      & $0.3$   & $5$     \\
$\beta_A$            & $-3$      & $0.8$   & $5$     \\
$c_\gamma$           & $0.05$    & $0.4$   & $3$     \\
$\alpha_\gamma$      & $-3$      & $0.0$   & $3$     \\
$\beta_\gamma$       & $-3$      & $0.0$   & $3$     \\
$c_{\lambda_v}$      & $10^{-6}$ & $0.001$ & $1$     \\
$\alpha_{\lambda_v}$ & $-3$      & $0.5$   & $5$     \\
$\beta_{\lambda_v}$  & $-3$      & $0.5$   & $5$     \\
\bottomrule
\end{tabular*}
\end{table}




\end{appendices}

\end{document}